# Event Memory and Narrative Dynamics: A Large Language Model Framework for Multidimensional State Spaces

Joshua Sariñana
Research Affiliate, Massachusetts Institute of Technology
Correspondence: jsarinan@mit.edu

ORCID: https://orcid.org/0000-0002-7452-569X

## Abstract

Studying event memory requires defining what constitutes an event and how events relate within remembered experience. Autobiographical narratives provide evidence of this organization, yet their analytical units must first be established. This study introduces a large language model framework for identifying events in narrative and modeling their representational structure and dynamics with excerpt-level traceability. Explicit boundary rules and a feature schema, informed by hippocampal models of episodic memory and navigation, guide event construction. In a proof-of-concept application to interviews from five participants, the framework constructed 256 event-states. Multiple correspondence analysis embedded these event-states in feature-based representational spaces. Analyses focused on the selected 4-D geometry, with 10-D and 13-D geometries used for higher-dimensional reference comparisons. Consecutive event-states were significantly closer in representational space than expected under randomized narrative order, indicating local continuity. Similarity between basin-state feature profiles corresponded with similarity between one-step routing profiles in the selected 4-D ten-basin solution. Within this solution, context, mechanism, and theme differentiated full-system pathway engagement. A successor representation modeled expected basin-state occupancy across successive steps. Correlations between basin-centroid distances and complete successor-profile distances were positive across all five basin solutions, with exploratory tests supporting correspondence in at least two solutions across all three geometries. Together, these findings connect feature-based event representations with observed narrative transitions and modeled multistep organization across multidimensional state spaces. By translating principles from hippocampal models into source-linked event boundaries and state representations, the framework provides an analytical basis for testing neural accounts of event segmentation and predictive organization during naturalistic recall.

## 1 Introduction

Cognitive systems can be analyzed through the similarity among states, the sequential relationships linking them, and the multistep structure that shapes which states are expected to be reached over longer sequences (Bellmund et al. 2018). Representational geometry characterizes how states are related according to similarity in their properties or response patterns (Diedrichsen and Kriegeskorte 2017; Greco et al. 2024; Shepard 1980). In sequential models, one-step dynamics are represented by probabilities over the next state conditional on the current state (Bellman 1957; Sutton and Barto 1981). The successor representation includes current-state identity and expected discounted occupancy across subsequent transitions, providing a model of longer-range relationships among states (Dayan 1993; Gershman 2018; Gershman et al. 2012). Computational models have also learned successor representations from simplified artificial-language sequences (Stoewer et al. 2022). Comparisons among these analytical descriptions require a shared set of states or explicit mappings between the state resolutions used in each analysis.

Analyzing these relationships requires a defined set of states, that is, analytically distinguishable configurations of represented content with properties and sequential relationships that can be measured. Depending on the domain, a state may correspond to a spatial location (O'Keefe and Dostrovsky 1971), an item–context representation (Ranganath 2010), a task configuration (Schuck et al. 2016), or an event (Baldassano et al. 2017; Zacks et al. 2007).

Whether representational geometry, transition structure, and successor structure preserve similar relationships among the same states provides a basis for comparing these forms of organization. If states that are close in representational geometry also have similar one-step transition distributions and complete successor profiles, this would indicate correspondence between representational and sequential organization. However, if the correspondences are incomplete, distinct analytical descriptions of representational geometry, transition structure, and successor structure may be required. Existing work suggests that partial correspondence is plausible. Temporal transition patterns can reshape learned representations without being reproduced exactly in the resulting geometry; contextual representations can bias retrieval order without determining it; and successor representations transform one-step transition structure into multistep relationships that are not reducible to immediate adjacency (Gershman 2018; Howard and Kahana 2002; Momennejad et al. 2017; Polyn, Norman, and Kahana 2009; Schapiro et al. 2013). Relatedly, distinct contextual memories encoded close in time recruit more overlapping neural ensembles than memories separated by longer intervals, showing that temporal proximity can influence how separate events are represented (Cai et al. 2016). The unresolved issue is the degree to which these forms of organization preserve corresponding relationships among the same states.

The ability to examine this correspondence also depends on how analytical states are defined and how their sequential relationships are established. In experiments organized around explicit state sequences, task design can specify both the analytical states and their possible transitions (Schuck et al. 2016). This control supports manipulation of the properties used to represent states and the transition structure linking them. However, jointly constraining these properties and transitions can introduce correspondence between representational geometry and one-step transition structure that partly reflects the experimental design. Naturalistic settings provide a complementary test when state identities, their represented properties, and their sequential relationships are not jointly prescribed by experimental design. Autobiographical narrative is particularly useful because remembered experience is organized into differentiated event content and narrated sequences (Cohn-Sheehy et al. 2021; Conway and Pleydell-Pearce 2000). However, the need to derive analytical states from narrative creates a model-construction problem. Event boundaries and state identities must be constructed or inferred from the narrative, with their order preserved, before representational geometry, one-step transition structure, and successor structure can be compared (Baldassano et al. 2017; Shin and DuBrow 2021).

A first step in resolving this model-construction problem is defining the event units that can serve as analytical states. Event-cognition research treats events as constructed representations rather than self-identifying units present in continuous experience (Baldwin and Kosie 2021; Kurby and Zacks 2008; Zacks et al. 2007). Event-segmentation and narrative-comprehension frameworks explain how experience is partitioned, represented, updated, and retrieved (Shin and DuBrow 2021; Zwaan, Langston, and Graesser 1995; Zwaan and Radvansky 1998). Recent experimental work further indicates that event segmentation can occur across predictable transitions, and transitions between stable contexts can produce stronger segmentation than isolated prediction errors (Güler, Serin, and Günseli 2025; Shim et al. 2024). Boundary identification alone does not establish the event representations needed to compare representational geometry with sequential organization. Autobiographical transcripts provide ordered evidence, but translating that evidence into analytical states requires an annotation system that specifies event boundaries and feature assignments while preserving their source passages and narrative order. Developing such a system requires deciding which properties of remembered experience distinguish and organize events, and how those properties can be expressed as explicit annotation rules.

Hippocampal-specific models of episodic memory emphasize that events are represented through relations among their context, spatial setting, temporal position, and content (Cohn-Sheehy et al. 2021; Eichenbaum 2017; Ranganath 2010; Sariñana et al. 2014). Models of hippocampal-dependent navigation provide an overlapping account of how spatial and relational information is organized in relation to locations, goals, reward, and trajectories (Behrens et al. 2018; Bellmund et al. 2018; Sariñana and Tonegawa 2016). Although episodic-memory and navigation models provide a structural basis for organizing remembered events, an event representation also requires characterizing the content being organized. Thematic structure provides a narrative-level account of what an event concerns, complementing the spatial, contextual, and relational information that situates it. Additionally, affect characterizes the evaluative significance attached to that content. Although affect is not a defining dimension of classic hippocampal accounts of spatial and episodic organization, emotional significance can influence hippocampal memory processing through interactions with broader affective-memory systems (Qasim et al. 2023; Richardson, Strange, and Dolan 2004). Thematic and affective features complement the contextual and relational organization supplied by episodic-memory and navigation models by characterizing what an event concerns and the valence associated with its content.

An additional methodological problem concerns whether representational and dynamical analyses require different levels of state resolution. Prior work has shown that continuous experience can be segmented at multiple scales and that event representations can support memory, prediction, and hierarchical organization (Baldassano et al. 2017; Franklin et al. 2020; Geerligs et al. 2022; Geerligs, Van Gerven, and Güçlü 2021). A related but distinct issue is representational dimensionality. More specifically, lower- and higher-dimensional representations can preserve different aspects of an experience, with lower-dimensional organization emphasizing shared or generalizable structure and higher-dimensional representations retaining finer distinctions (Kerrén et al. 2025). Related work on state abstraction has examined how fine-grained states can be compressed into coarser representations for learning and prediction, often by grouping states according to similarities in their transition, reward, or value structure (Castro 2020; Ferns, Panangaden, and Precup 2004). Applying state abstraction to autobiographical narratives first requires deriving analytical states from the observations and preserving explicit mappings as those states are aggregated. This traceability is important because changing state resolution can itself alter the relationships being measured. Preserving the mapping across resolutions makes it possible to test whether a relationship between representational and transition structure is maintained as the same underlying observations are transformed, or whether it changes with analytical resolution.

The present study develops an annotation system for constructing source-linked event-states from autobiographical narrative and a quantitative framework for examining their representational and sequential organization. The annotation system operationalizes broad principles from hippocampal-dependent episodic-memory and navigation models as an event-structure schema and explicit boundary rules, complemented by thematic and affective annotation procedures. Schema-constrained large language model (LLM) procedures apply these specifications to narrative language. Preserved excerpt identities and excerpt-to-event mappings link the resulting annotations to bounded events and their source evidence. Each event-state identifies a particular occurrence in a participant's narrative sequence. The selected categorical event-state features define the representational geometry, while preserved within-participant narrative order supplies observed event-state adjacencies. Macro-clusters characterize broad feature organization. Basin-states pool nearby event-states to provide repeated outgoing observations for estimating one-step destination distributions, represented as routing profiles. Basin centroids summarize the mean positions of constituent event-states, while basin-state feature profiles summarize their categorical composition. Explicit mappings link source excerpts to event-states and link those event-states separately to basin-states and macro-clusters, preserving traceability across analytical resolutions.

A proof-of-concept application to semi-structured autobiographical interviews from five participants uses the constructed event-states to examine how representational organization relates to narrative progression. Event-level analyses assess local feature continuity and displacement patterns between consecutive event-states. Macro-cluster analyses compare between-cluster transition probabilities with distances between macro-cluster centroids and examine how current macro-cluster membership relates to the next basin-state. Feature–routing correspondence tests whether similarities between basin-state feature profiles correspond with similarities between their routing profiles. Feature-conditioned comparisons examine whether primary context, mechanism presence, and event-level meta-theme are associated with different pathways across the basin-state system or with different destination distributions from the same adequately supported source basin-state. Representational–successor analyses compare pairwise basin-centroid distances with corresponding distances between complete successor profiles, extending the comparison to modeled occupancy across repeated basin-state transitions. Comparisons across representational geometries and basin solutions assess how these relationships depend on dimensionality and aggregation.

The present interviews include narratives of creative practice and community participation. Constructing event-states from these narratives requires specifying which distinctions in the described activities are represented and which define separate events. The annotation schema determines which properties become available for computational comparison, while source mappings preserve the narrative evidence supporting the resulting representations. Making these decisions explicit connects AI-assisted narrative analysis with digital humanities research on how cultural experience is represented through computational methods.

## 2 Methods

### *2.1 Participants, recruitment, and interviews*

Eight adults were recruited through purposive and referral sampling; five completed the full semi-structured autobiographical interview and comprise the analytical sample. Recruitment included local artists and other creative practitioners identified through Massachusetts Cultural Council grant listings, followed by participant referrals. The analytical sample included one male and four female participants; age was not collected. Introductory interviews were used to obtain background information and inform the full-interview questions but were not included in the analyses.

Full interviews were conducted by the author using a videoconferencing platform, recorded, and automatically transcribed. Only interviewee speech was analyzed. Questions and follow-up probes elicited autobiographical narrative concerning developmental and geographic background, community membership, spatial navigation, creative and professional activity, digital-media use, and interactions with their environments. Interviews lasted 98–205 min (total, 11 h 25 min; median, 2 h 20 min), see Table S1B.

The interviews were conducted as part of an independent art and documentary project. The author did not have access to an institutional ethics-review process for this project. No formal institutional review board review or exemption determination was obtained. Participants were informed of this status and provided written consent for recorded interviews, computational analysis using machine-learning systems, and use of interview quotations and derived materials in artistic and scholarly outputs, including academic publications. The public analytical dataset was de-identified using replacement participant and event identifiers. Original identifiers, recordings, and complete transcripts are withheld from public distribution to protect participant privacy. The consent template is available upon request.

### *2.2 Narrative excerpt and event-state construction*

Interviewee autobiographical narrative text was analyzed using two predefined schema-constrained prompts developed for this study, one for thematic analysis and one for hippocampal-informed event-structure analysis. The thematic prompt identified meaning-bearing passages and retained them as ordered verbatim excerpts. Each excerpt received superordinate-theme, sub-theme, thematic-type, and sentiment assignments. After superordinate-theme labels had been collected across all participants, they were grouped by semantic meaning into a shared 12-category meta-theme ontology. The resulting mapping was applied to excerpt records by deterministic lookup (Supplement S2). The hippocampal-informed event-structure prompt analyzed the same narrative text using the identified excerpts as source units and grouped consecutive excerpts into bounded events according to the predefined event-structure schema and boundary rules. For LLM inference, each participant's transcript was divided into processing segments of approximately 10 pages to maintain manageable input length and consistent inference quality. The full participant transcripts averaged 38 pages. The complete analytical workflow is shown in Fig. 1.

Initial segment-level thematic and hippocampal-informed event-structure annotations were generated with the ChatGPT-5.1 Thinking Extended model and integrated using predefined prompts with GPT-5.1 Pro. Claude Opus 4.7 was used only to audit source support, internal consistency, and schema adherence; it did not generate or replace final assignments. No independent human coding of event boundaries or feature assignments was used. The author separately assessed excerpt-level source correspondence through the manual audit described in Section 2.8. Event-construction constraints, integration procedures, and audit checks are detailed in Supplements S1–S2. The annotation codebook and model-independent annotation-constraint prompt templates are included in the public reproducibility package (Supplement S7; repository details are provided in the Data Availability statement).

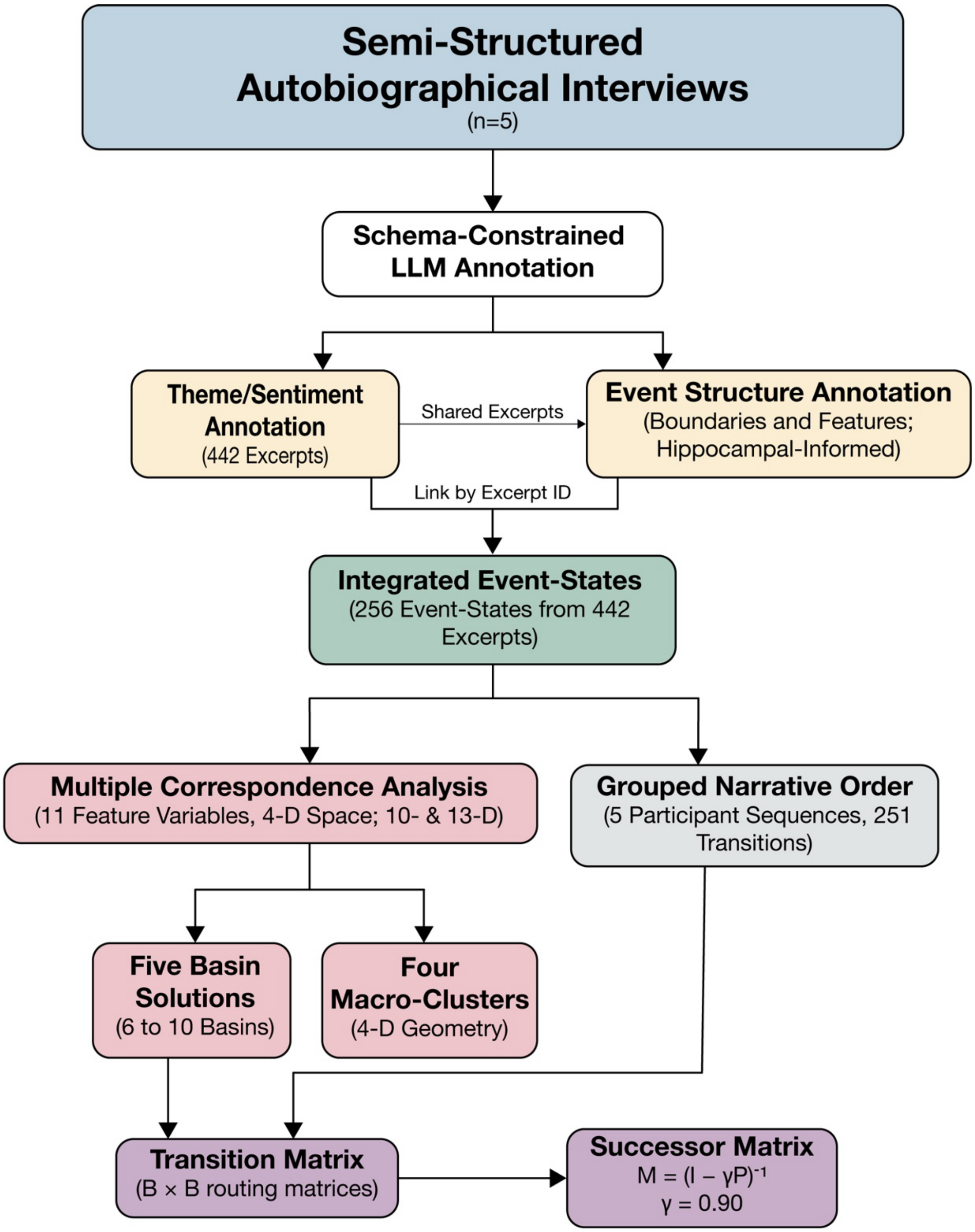

**Figure 1. Construction and analysis of narrative-derived state spaces.** Schema-constrained thematic and event-structure annotation of interviews from five participants produced 442 excerpts and 256 source-linked event-states. Eleven active categorical variables defined the representational geometry through multiple correspondence analysis (MCA). Four macro-clusters summarized broad organization in the 4-D macro-geometry. Separately, event-states were assigned to five basin solutions: ten, nine, and six basin-states fitted in 4-D, and nine and eight basin-states fitted in the two-dimensional Dim0 × Dim3 projection. Within-participant narrative order supplied 251 event-state adjacencies. Pooling these adjacencies by basin assignment defined each solution's one-step transition matrix P, from which the complete successor matrix M was calculated at $\gamma = 0.90$. Complete successor profiles include current-state identity and expected discounted occupancy across successive basin-state transitions. The 10-D participant-robust geometry and 13-D full retained geometry provided higher-dimensional reference comparisons.

Each bounded event was characterized by categorical fields describing hippocampal-informed mechanism categories, context, anchor category, landmark class, orientation, perspective, reference frame, and representational domain, together with a free-text specific anchor. The specific anchor captured the fine-grained referential focus organizing an event and was not treated as a persistent entity identifier across the interview. A new event began when the dominant mechanism category, primary context, specific anchor, or anchor category changed. Representational domain could also contribute to a boundary when the dominant representational mode changed across at least one complete excerpt; it did not independently define any realized boundary in the final dataset. Feature roles and boundary rules are described in Supplement S1; the active MCA variables are listed in Table S1A. Complete category definitions and assignment rules are provided in the public annotation codebook.

The integrated thematic and hippocampal-informed event-structure outputs linked 442 ordered excerpts to 256 event-states through preserved excerpt identifiers and excerpt-to-event mappings. This join associated each bounded event with its event-structure features and with the thematic annotations of its constituent excerpts; the thematic variables used downstream were represented at the event level according to the integration procedure described in Supplement S2. The resulting analysis table contained one row for each of the 256 event-states and 20 columns, comprising 11 active MCA variables and nine identifier, sequence, and descriptive fields (Table S1A; repository workbook field inventory, Supplement S7). The public reproducibility package uses a separate de-identified, 19-field event-level input table described in Supplement S7.

### *2.3 State-space construction*

The event-level table was used to construct the representational geometry with multiple correspondence analysis (MCA) (Greenacre and Blasius 2006). Eleven categorical variables entered the MCA, comprising primary and secondary mechanism, primary and secondary context, anchor category, landmark class, orientation, perspective, reference frame, event-level meta-theme, and event-level sentiment. The specific anchor and representational domain contributed to event construction but were not active MCA variables.

The complete MCA solution contained 54 dimensions. Parallel analysis tested how many dimensions captured feature relationships exceeding those expected after randomizing feature assignments within participants (Horn 1965). Using 5,000 permutations, this procedure retained 13 dimensions. Repeating MCA and parallel analysis after excluding each participant retained at least 10 dimensions in every comparison. These results defined the 13-D full retained geometry and the 10-D participant-robust geometry.

A separate k-means comparison evaluated two through six macro-clusters across two through 13 dimensions to identify a compact representation of broad feature relationships. Each macro-cluster was required to contain at least 5% of event-states and represent at least two participants. Diagnostics assessed cluster separation and whether event-states remained assigned together after resampling events or excluding participants. Additional checks omitted predefined sets of related feature variables from MCA and clustering, then tested whether the resulting macro-clusters differed in the excluded features (Supplement S3). Four dimensions were the smallest tested representation supporting four macro-clusters that met the population requirements. Compared with the four-cluster solutions in 10-D and 13-D, this solution also had clearer separation and more consistent assignments. The 4-D geometry was selected for the principal analyses, while the 10-D and 13-D geometries supplied reference comparisons retaining more of the complete MCA distance structure. Procedures and diagnostics are reported in Supplement S3, Tables S2A–D, and Figure S1; complete comparisons are generated by the public reproducibility workflow (Supplement S7).

### *2.4 Basin-state construction and multiscale mapping*

Basin-states pooled nearby event-states to provide repeated outgoing observations for estimating transition probabilities. Solutions were fitted either in the first four MCA dimensions or directly in the selected Dim0 × Dim3 projection. This comparison assessed how basin assignments and downstream relationships changed when aggregation used all four coordinates or only the two displayed coordinates. The dimensional prefix in each basin-solution name identifies its fitting space, independently of the geometry used to measure subsequent distances (Table 1).

The display plane was selected primarily by whether event-states appearing nearby in the projection were also nearby in the reference geometry, measured using neighborhood trustworthiness (Venna and Kaski 2001). Preservation of pairwise distances, assessed using Spearman and Pearson correlations, supplied secondary criteria. The selected Dim0 × Dim3 plane was used to display macro-clusters, event-level displacement fields, and basin-states. It also supplied the fitting coordinates for the 2-D basin solutions. For the 4-D solutions, Dim1 and Dim2 could influence membership despite being absent from the display. Projection comparisons are reported in Table S2B and Supplement S3.

| **Property** | **4-D solutions** | **2-D solutions** |
|---|---|---|
| Coordinates used for basin fitting | Dim0, Dim1, Dim2, Dim3 | Dim0 and Dim3 |
| Retained basin counts | 10 (selected), 9, 6 | 9 (comparison), 8 |
| Projection | Dim0 × Dim3 | Dim0 × Dim3 |
| Primary centroid-distance geometry | 4-D | 4-D |

Table 1. Basin fitting, distance measurement, and display spaces. Dimensional prefixes identify basin-fitting coordinates. All solutions share the Dim0 × Dim3 display and use 4-D centroid distances for the primary distance comparisons.

Size-constrained k-means assigned event-states to basin-states by minimizing their squared Euclidean distances from the assigned centroids while enforcing lower and upper event-count bounds. The screen examined 5–20 basin-states, lower bounds of 8, 9, or 10 event-states, and upper bounds of 25–50, with at least eight outgoing observations required per basin-state. Initial screening considered geometric separation and transition support. Repeated fits from different starting centroids checked sensitivity to initialization. Event omissions, including analyses that refitted MCA, assessed whether similar basin assignments and sufficient outgoing observations were retained after changes to the sample (Lange et al. 2004). Assignment agreement was measured using the adjusted Rand index (ARI) (Hubert and Arabie 1985). The staged selection procedure excluded near-duplicate partitions and retained up to three 4-D and two 2-D solutions before routing and successor comparisons (Supplement S4; Table S2G; Fig. S3).

The selected 4-D ten-basin solution, with event-count bounds of 9–47, ranked first among eligible 4-D candidates under the combined assignment-stability and outgoing-support criteria. The 2-D nine-basin solution, with bounds of 8–50, supplied the principal comparison fitted directly in Dim0 × Dim3. The retained 4-D nine-basin, 4-D six-basin, and 2-D eight-basin solutions supplied additional supported aggregations for assessing dependence on basin construction (Table S2E). Because basin counts and event-count bounds also differed, these comparisons did not isolate the effect of fitting dimensionality.

All five solutions reassigned the same 256 event-states while preserving their 251 narrative adjacencies, allowing changes in the measured relationships to be attributed to aggregation of the same observations. Macro-clusters were fitted separately from basin-states, with both retaining mappings to their constituent event-states. Each basin centroid was the mean position of its assigned event-states in the MCA coordinates used for the distance comparison. For every basin solution, centroid distances were measured primarily in the 4-D geometry; the 10-D and 13-D comparisons changed the distance geometry while retaining the same memberships and successor profiles.

## *2.5 Example of state construction and transition aggregation*

Table 2 traces two consecutive events from their constituent excerpts to their event-state representations and basin assignments. The first event comprised two consecutive excerpts and was assigned to B2; the following event comprised five consecutive excerpts and was assigned to B1. The boundary coincided with changes in the specific anchor, primary mechanism, and primary context, while anchor category remained a static landmark. This example illustrates the fine-grained role of the specific anchor. The referential focus changed from the Newberry space to the Newberry exhibition even though the coarser anchor category did not. Because the events were consecutive in narrated recall order, their assignments contributed one observed B2 → B1 transition to the pooled basin-state transition count matrix.

| **Attribute** | **First event** | **Following event** |
|---|---|---|
| Number of excerpts | 2 | 5 |
| Representative passage | "we agree to do an event and set a budget on what things were gonna cost out of pocket and ... set a goal of attendance and what we hope to achieve with this." | "I showed my ... projection mapping. ... my brainwave activity and drawings ... some sculptures ... performance pieces ... boxing and ballet." |
| Bounded event | Negotiating the terms, resources, and intended outcomes of an exhibition | Describing the works assembled within the resulting exhibition |
| Primary mechanism | Goal-directed navigation | Context integration |
| Primary context | Social | Physical |
| Specific anchor | Newberry space | Newberry exhibition |
| Anchor category | Static landmark | Static landmark |
| Meta-theme / sentiment | Institutions, Gatekeeping, and Resource Distribution / Positive | Institutions, Gatekeeping, and Resource Distribution / Neutral |
| Basin-state assignment | B2 | B1 |

Table 2. Worked example tracing excerpt-level event construction to basin-level transition aggregation. Representative passages are source-verbatim selections from the excerpts constituting each event; ellipses indicate omitted words or intervening material.

## *2.6 Transition analyses*

The 256 event-states followed each participant's narrative order, yielding 251 consecutive transitions. For each basin solution, these event transitions were mapped to their source and destination basin-states and pooled across participants, including self-transitions. Dividing each basin-state's destination counts by its total outgoing count produced the empirical transition matrix

P without smoothing. Each row defined a routing profile containing the probabilities of its possible next basin-states.

Two exploratory event-level analyses tested whether local progression differed from randomized narrative order. The event-distance analysis asked whether consecutive event-states were closer in the 4-D geometry than expected after shuffling. The displacement-field analysis asked whether the average direction and magnitude of successive movements in Dim0 × Dim3 differed from their shuffled expectation. Both randomized interior event order within participants while fixing the first and last event-states (Supplement S5; Tables S3D–E; Fig. 2B–D). Separate fields grouped transitions by the source event's primary context or anchor category to describe how projected displacement varied with these features (Fig. S2).

Macro-cluster analyses tested whether between-cluster transition probabilities varied with centroid distance. A complementary analysis asked whether knowing the current macro-cluster provided information about the next basin-state beyond that expected under randomized narrative order, using conditional mutual information calculated within participants (Cover and Thomas 2006). A descriptive participant-exclusion comparison then assessed whether adding current basin-state identity improved next-basin prediction beyond current macro-cluster identity. Predictions for each excluded participant used transition probabilities estimated from the other participants, while geometry and memberships remained fixed (Supplement S5; Tables S3A and S3E–F).

Feature–routing correspondence tested whether basin-states with similar feature composition had similar one-step destination distributions. Pearson correlations compared pairwise feature-profile cosine similarities with routing-profile cosine similarities. Primary feature profiles contained category counts for mechanism presence, primary context, and event-level meta-theme. Mechanism presence combined primary and secondary assignments, counting each distinct mechanism once per event. Comparisons were repeated across all five basin solutions and using all 11 active MCA variables (Table S3B; Fig. S4).

Feature-conditioned analyses asked whether source-event features were associated with different pathways across the basin-state system or with different destinations from the same source basin-state. Conditions required at least 10 source transitions. Comparisons from a shared source required at least three outgoing transitions under each condition. Feature labels were shuffled across pooled source transitions to test whether routing differed between conditions more than expected under random label assignment. The same support requirements were reapplied after every shuffle. Benjamini–Hochberg correction controlled the false discovery rate across the three feature comparisons separately within each basin solution and comparison type (Benjamini and Hochberg 1995). Results are reported in Table S3C, with detailed condition-support tables in the public reproducibility package (Supplement S7).

Holm correction controlled the chance of a false-positive result across the pooled displacement-field test and the five tests relating current macro-cluster membership to the next basin-state. The event-distance analysis was evaluated separately. Macro-cluster distance–transition and feature–routing comparisons reported nominal p-values without adjustment across those comparisons. Supplement S5 provides the test statistics, randomization procedures, and prediction-scoring methods.

### *2.7 Successor structure*

One-step transition profiles describe immediate destinations. The standard successor representation transforms the complete transition matrix into profiles of current-state identity and expected discounted occupancy across repeated transitions (Dayan 1993; Gershman 2018).

$$M = (I - \gamma P)^{-1}$$

with $\gamma = 0.90$, giving progressively less weight to occupancy at later steps. Each row of M represented one basin-state by its current-state identity and expected discounted occupancy of all basin-states across repeated transitions.

Representational–successor correspondence tested whether basin-states closer in the 4-D geometry also had more similar complete successor profiles. For each of the five basin solutions, Pearson correlation compared pairwise Euclidean basin-centroid distances with distances between complete successor profiles. Two-sided permutation tests assessed whether the observed correlation was more extreme than expected under arbitrary matching of basin identities. Each permutation rearranged the rows and columns of the successor-distance matrix together while leaving centroid distances fixed, preserving the relationships within each matrix (Mantel 1967). Reference comparisons used 10-D and 13-D centroid distances to assess whether correspondence extended to the higher-dimensional geometries. Individual comparisons reported nominal p-values; the across-solution procedures described below addressed the specified multiple-comparison claims (Supplement S6; Table S4A).

For visualization, principal component analysis reduced complete successor profiles to two dimensions. Matched two-dimensional Procrustes analysis (Gower 1966) compared this projection with basin centroids in Dim0 × Dim3 (Table S4B; Figs. 5A and S5A–C). Complete successor profiles supplied the distance-correspondence analyses. Sensitivity checks examined discount factors, participant omissions, basin-visit transitions, profile definitions, and narrative order (Tables S4C–F; Fig. S6; repository discount-factor sweep, Supplement S7). Order controls rebuilt transition and successor matrices after unrestricted within-participant shuffling.

Partial-conjunction tests assessed whether correspondence was supported in at least two, and subsequently at least three, of the five basin solutions while allowing dependence between solutions (Wang and Owen 2019). Holm adjustment identified which individual solutions retained statistical support after accounting for the five comparisons (Holm 1979). A stricter assessment required support across all three distance geometries within the same basin solutions (Tables S4G–H). A separate joint narrative-order test compared the median correlation across all five solutions with shared within-participant order shuffles, preserving dependence between solutions (Simonsohn, Simmons, and Nelson 2020; Winkler et al. 2016). The largest joint p-value across geometries assessed whether the aggregate result was supported in every distance space (Table S4I). Supplement S6 provides the complete procedures.

### *2.8 Source traceability and reproducibility*

Technical checks assessed required fields, permitted categorical values, identifier consistency, and excerpt-to-event mappings. These mappings preserve the source provenance of each constructed event-state.

The author manually audited a participant-stratified random sample of 150 of the 442 excerpts, with sample sizes allocated proportionally across participants. Excerpts were sampled without replacement and compared against the original interview transcripts. An excerpt was classified as a failure if it could not be located in its source transcript or differed beyond trivial formatting changes. An exact, conservative one-sided 95% upper confidence bound on the dataset-wide failure proportion accounted for the participant-specific population and sample sizes (Supplement S1; Table S1D).

The public reproducibility package supports numerical reproduction from frozen event-level annotations; it does not regenerate the upstream model-based annotations. Package contents and reproduction procedures are detailed in Supplement S7, with repository information provided in the Data Availability and Code Availability statements.

### *2.9 AI-assisted manuscript preparation*

Generative AI tools were used to support several stages of this study. Large language models were used for schema-constrained analysis and annotation of interview transcripts, including construction of structured event-level representations. LLMs also assisted in generating Python scripts for statistical analyses and figure production, with GPT-6 used for script development and auditing. All downstream computational analyses of the transcript-derived representations were executed locally. Quantitative figure panels were rendered locally from analysis outputs using plotting scripts; no generative image tools were used. The author reviewed the scripts and inspected their outputs. During manuscript preparation, LLMs were used to assist with editing and to review the text for consistency with the analyses and reported results. The models used were ChatGPT-5.1 Thinking Extended, GPT-5.1 Pro, GPT-6, and Claude Opus 4.7. The author reviewed and approved all code, analyses, interpretations, and final manuscript content.

## 3 Results

### *3.1 Event construction preserves narrative sequence and source links*

The annotation procedure converted 442 ordered excerpts into 256 source-linked event-states while preserving 251 consecutive transitions within participant narratives. All event-states entered the representational and transition analyses. The excerpt-to-event mappings retained the narrative evidence underlying each analytical unit (Table S1B).

All 150 audited excerpts were located in their corresponding source transcripts and matched the source wording apart from trivial formatting differences. No traceability failures were observed. The one-sided 95% upper confidence bound was seven failures among the 442 excerpts, equivalent to 1.58% (Supplement S1; Table S1D).

Changes in specific referential focus characterized nearly every within-segment event boundary. The specific anchor changed at 234 of 235 boundaries, usually alongside changes in other event features. At 40 boundaries, it was the only tracked feature to change, allowing events to be distinguished within otherwise stable categorical descriptions. These counts describe the realized boundaries; the complete feature-change summary is reported in Table S1C. Event construction retained distinctions in referential focus alongside categorical feature descriptions, providing source-linked analytical units for comparing representational organization with narrative sequence. Preserved event-state identities and narrative adjacencies also allowed alternative basin solutions to be compared using the same underlying narrative observations.

### *3.2 Four dimensions resolve four supported macro-clusters*

The 4-D geometry was the smallest tested representation that resolved four macro-clusters meeting the event-count and participant-support criteria. Compared with the four-cluster solutions in 10-D and 13-D, it provided greater geometric separation and higher assignment stability, with similar held-out feature alignment (Supplement S3; Tables S2A–C; Fig. S1).

The four macro-clusters contained 133, 59, 17, and 47 event-states, with at least four participants represented in each (Fig. 2A). Their post hoc feature profiles distinguished social context integration, prospective and goal-related organization, virtual-digital mediation, and physical-spatial place/path organization. When related feature variables were omitted from MCA and clustering, the resulting macro-clusters differed in the excluded features in five of the six comparisons after correction ($q < 0.001$; Table S2D). Dim0 × Dim3 supplied the selected projection of the 4-D geometry; full coordinates were retained for the corresponding clustering and distance analyses (Table S2B). These findings supported a compact 4-D representation of broad feature organization, providing a common geometry for examining how narrative transitions relate to event-state positions.

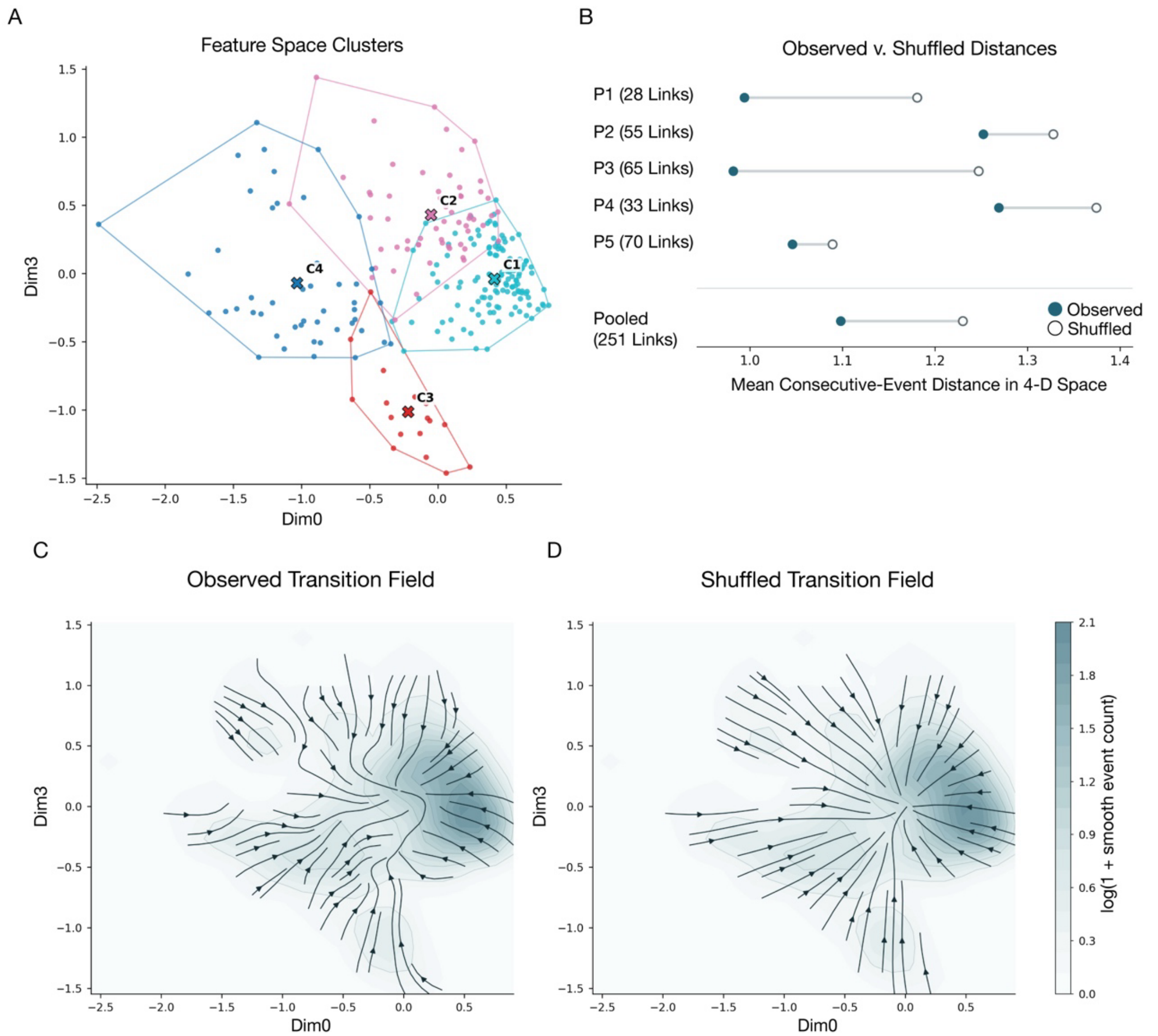

**Figure 2. Broad representational organization and local narrative transitions. (A)** The 256 event-states displayed in the Dim0 × Dim3 projection, colored by four macro-clusters fitted in the 4-D geometry. Macro-clusters C1–C4 contain 133, 59, 17, and 47 event-states, respectively. Crosses mark projected centroids; outlines enclose projected cluster members. **(B)** Observed and expected mean Euclidean distances between consecutive event-states in the complete 4-D macro-geometry, shown for each participant and pooled across 251 adjacencies. Filled circles show observed means; open circles show exact expectations under randomized interior narrative order within participants, preserving each participant's first and last event-states. Connecting lines join the paired estimates. **(C)** Observed mean one-step displacement field in the Dim0 × Dim3 projection, calculated from the 251 within-participant event-state adjacencies. Streamlines follow the locally averaged displacement from source to destination event-states, with arrows indicating direction. They summarize the estimated field and do not represent observed multistep trajectories. **(D)** Exact expected mean displacement field under the same endpoint-fixed narrative-order shuffling used in B. This field represents the expectation across permissible randomized orders, rather than one shuffled sequence. Panels C and D use identical coordinates, gridding, smoothing, and support thresholds.

### *3.3 Narrative transitions preserve local feature continuity*

Consecutive event-states were 10.7% closer in the 4-D geometry than expected under endpoint-fixed within-participant narrative-order shuffling ($p = 1 \times 10^{-6}$). Observed mean distances were below their shuffled expectations in all five participants (Fig. 2B). The pooled projected mean displacement field also departed from its shuffled-order expectation (Holm-adjusted p = 0.007; Fig. 2C–D).

At the four-cluster macro scale, centroid distance did not reliably correspond with immediate between-cluster transition probability (Table S3A). Additional macro-to-next-basin dependence and held-out prediction analyses are reported in Supplement S5 and Tables S3E–F. The pooled field result concerns the global displacement pattern; individual arrows and context- or anchor-conditioned fields remain descriptive (Fig. S2; Supplement S5). The event-level findings show local feature continuity and an average displacement pattern that differs from randomized narrative order. Basin-state analyses next examined whether similar feature profiles also corresponded to similar one-step destination distributions.

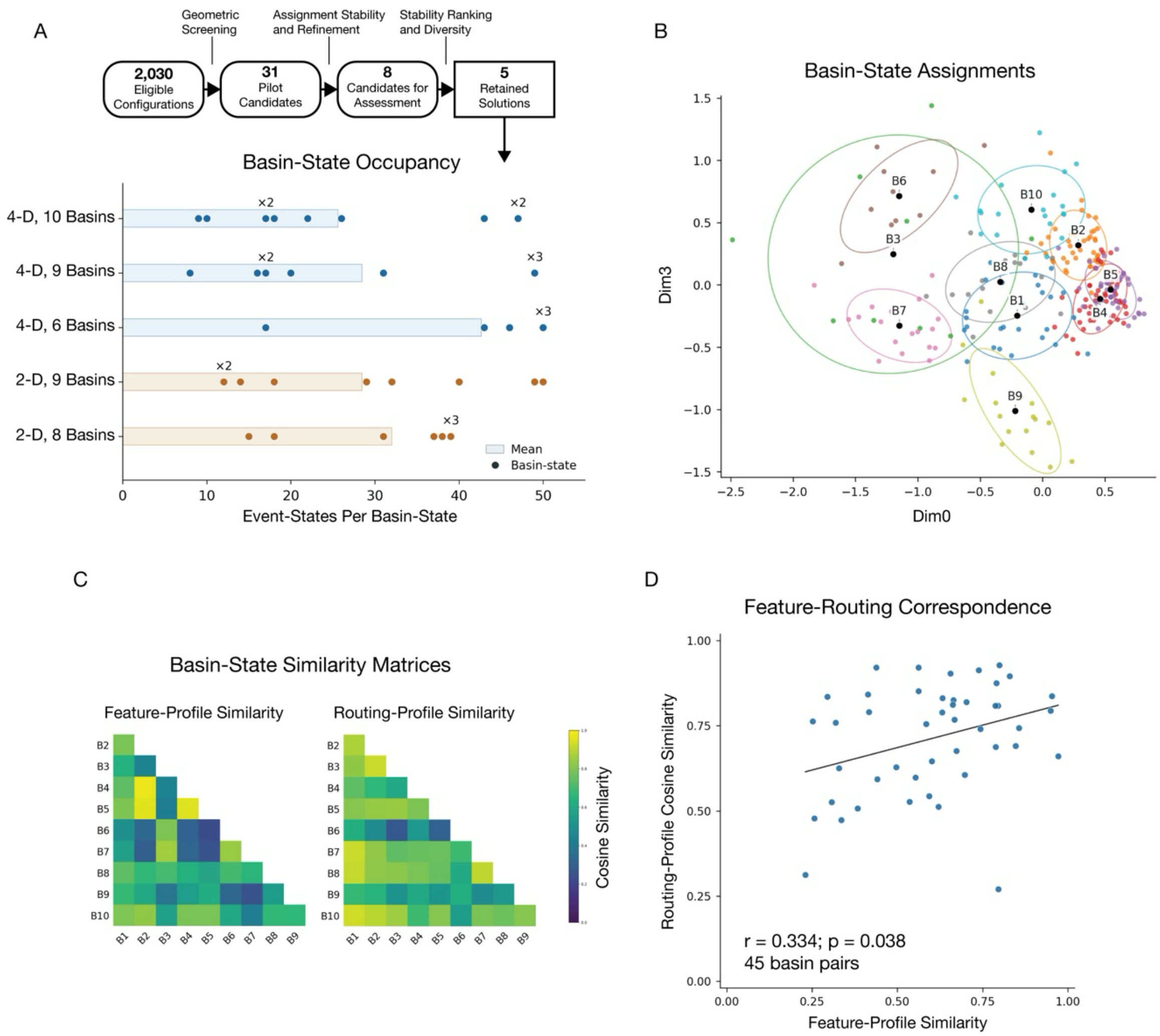

**Figure 3. Basin-state construction and feature–routing correspondence. (A)** Operating-resolution selection and occupancy of the five retained basin solutions. Geometric screening reduced 2,030 eligible construction configurations to 31 pilot candidates; eight advanced to reserved omission assessments, from which five solutions were retained. Different configurations could yield identical event-state partitions. Retention was capped at three 4-D and two 2-D solutions. Horizontal bars show mean event-state count per basin-state. Circles show individual basin-state counts; ×n identifies coincident counts. **(B)** The selected 4-D ten-basin solution projected onto Dim0 × Dim3. Colored points represent the 256 event-states, and black circles mark basin centroids. Ellipses summarize projected spread at 1.5 standard deviations. Basin membership was determined in the complete 4-D fitting space. **(C)** Pairwise cosine similarities between basin-state feature profiles and between routing profiles. Feature profiles summarize mechanism presence, primary context, and event-level meta-theme. Mechanism presence combines primary and secondary assignments, counting each distinct mechanism once per event-state. Routing profiles contain one-step transition probabilities to all ten basin-states, including the source basin-state. **(D)** Correspondence between feature-profile similarity and routing-profile similarity across 45 unordered basin pairs in the selected solution. Each point represents one basin pair.

### *3.4 Similar basin-state features correspond to similar routing profiles*

Pooling nearby event-states into the selected 4-D ten-basin solution permitted comparison of feature composition with one-step destination distributions (Figs. 3A and 4A). Basin-states contained 9–47 event-states, each with at least nine outgoing transitions. Support and geometric separation varied across basin-states (Tables S2E–F; Figs. 3A and S3B).

Basin-states with more similar profiles of mechanism presence, primary context, and event-level meta-theme had more similar routing profiles in the selected solution ($r = 0.334$, nominal $p = 0.038$; Fig. 3C–D). The association was positive across all five solutions and nominally significant in four (Table 3). Using all 11 active variables also produced positive associations across all five solutions, with nominal significance in two; the selected solution's estimate was $r = 0.290$, nominal $p = 0.053$ (Table S3B). Participant-omission checks also varied in statistical support (Supplement S5). These results connect basin-state feature composition with one-step destination distributions, with support depending on feature definition and basin construction.

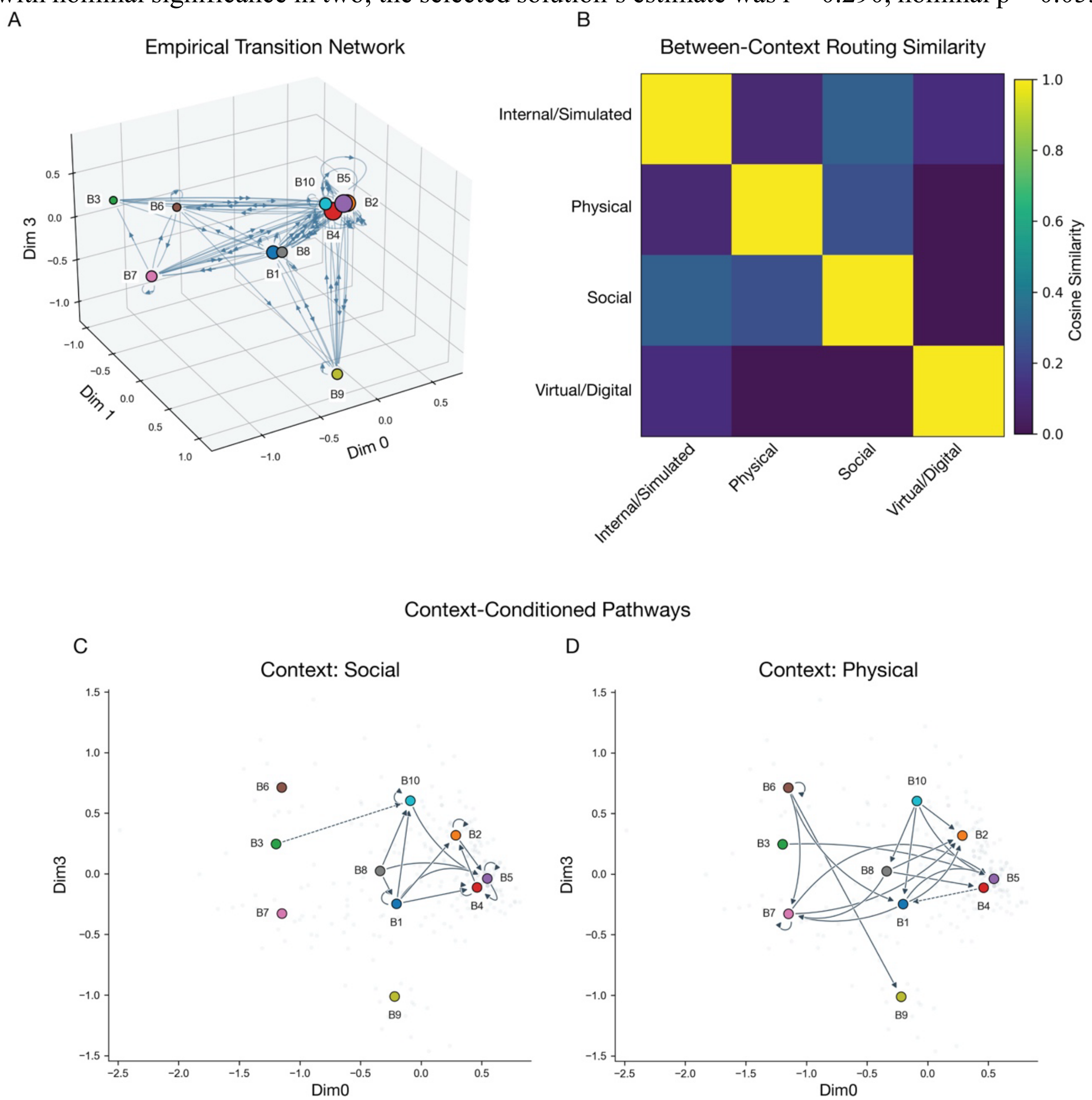

**Figure 4. Observed basin connections and context-conditioned pathway engagement. (A)** All observed directed connections displayed at basin centroids in Dim0, Dim1, and Dim3. Circle area represents event-state count. Arrows indicate transition direction; loops indicate consecutive event-states assigned to the same basin-state. **(B)** Pairwise cosine similarities between complete routing matrices conditioned on the primary context of the source event-state. Mean between-context similarity was lower than expected under shuffled condition labels ($q < 0.001$), supporting context-dependent full-system pathway engagement. The q-value is Benjamini–Hochberg adjusted across context, mechanism presence, and meta-theme within this basin solution and test family; it applies to the aggregate comparison, not to individual heatmap cells. Comparisons restricted to shared, adequately sampled source basin-states are reported in Table S3C. **(C-D)** Selected connections from social-context source event-states (left; 143 transitions) and physical-context source event-states (right; 64 transitions), displayed in Dim0 × Dim3. Each source shows its most frequent observed non-self destination, including ties, and all observed self-loops. Dashed arrows indicate one or two outgoing transitions from the source under that context; solid arrows indicate at least three. Circles mark basin centroids, and faint dots show event-state positions.

### 3.5 Feature conditions differentiate full-system pathway engagement

Primary context, mechanism presence, and meta-theme differentiated full-system pathway engagement in the selected 4-D ten-basin solution. Conditioned transition matrices were less similar than expected under shuffled condition labels, with corrected support for context and mechanism presence (both $q < 0.001$) and meta-theme ($q = 0.031$). Context and mechanism effects were supported across all five basin solutions; meta-theme effects were supported in the three 4-D solutions (Table S3C). Figure 4B–D shows context-conditioned comparisons and example pathways. Full-system and supported-source comparisons for all three feature comparisons are reported in Table S3C, with detailed support summaries in the package (Supplement S7).

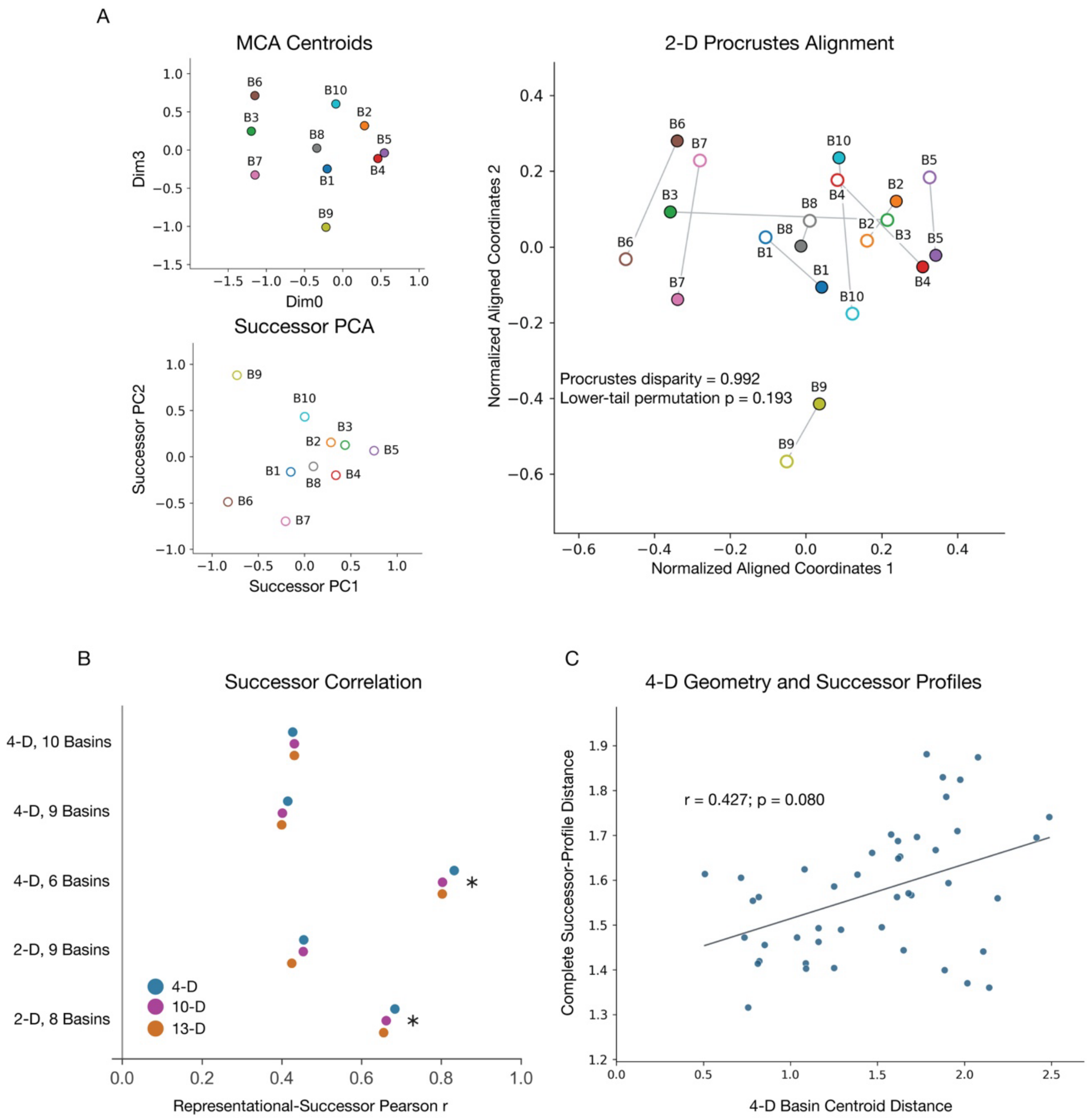

**Figure 5. Representational–successor correspondence across basin solutions.** **(A)** Basin centroids in the Dim0 × Dim3 representational projection (upper left), complete successor profiles projected onto Successor PC1 × PC2 (lower left), and the two-dimensional Procrustes alignment of these displays (right). The two successor components retain 36.1% of successor-profile variance. In the alignment, filled circles show normalized representational positions; open circles show aligned successor positions. Colors and labels identify corresponding basin-states. Connecting lines indicate alignment residuals. **(B)** Pearson correlations between pairwise basin-centroid distances and distances between complete successor profiles across all five basin solutions. Blue, purple, and orange circles indicate centroid distances calculated in the 4-D, 10-D, and 13-D geometries, respectively. The dimension in each row label identifies the geometry used to construct that basin solution. Circles are offset vertically for visibility. A single asterisk applies to all three circles in its row, identifying correspondence supported across all three distance geometries after Holm correction across the five solutions. The supported solutions are 4-D 6-Basin and 2-D 8-Basin. Rows identify the basin-fitting space and basin count. Colors indicate the geometry used to measure basin-centroid distances, with basin membership and successor profiles held fixed within each row. **(C)** Pairwise Euclidean distances between basin centroids in the complete 4-D geometry versus distances between complete successor profiles for the main solution, across 45 basin pairs. The successor analyses use $\gamma = 0.90$.

Evidence for different destinations from the same adequately sampled source basin-state was less consistent. Supported-source comparisons were nonsignificant in the selected solution, and the mechanism-related result in the 2-D nine-basin solution depended on participant composition. Features were associated with which pathways were engaged across the basin-state system, while destination differences from shared source basin-states remained unresolved (Supplement S5; Table S3C).

### *3.6 Representational–successor correspondence extends across basin solutions*

Basin-centroid distances were positively associated with complete successor-profile distances across all five basin solutions. Exploratory partial-conjunction tests supported correspondence in at least two solutions using 4-D distances ($p = 0.022$), and when correspondence was required across 4-D, 10-D, and 13-D distances within each solution ($p = 0.044$). Holm-adjusted tests identified the 4-D six-basin and 2-D eight-basin solutions as supported across all three geometries (Fig. 5B; Table 3; Tables S4G–H).

The selected 4-D ten-basin estimate was positive but nonsignificant ($r = 0.427$, nominal $p = 0.080$; Fig. 5C). The estimates describe correspondence across alternative aggregations of the same event-states, with statistical support depending on basin construction. These dependent solutions do not constitute independent replications. Figure 5A shows the projected representational and successor positions and their alignment for the selected solution; Figure 5C shows its complete-profile distance comparison. Corresponding displays for the 2-D nine-basin solution are provided in Figure S5.

Sensitivity analyses examined distance geometry, transition unit, and profile definition. Joint narrative-order evidence varied across geometries, and profile controls did not isolate a unique contribution from accumulated multistep structure. The estimates and participant-omission summaries are reported in Supplement S6 (Tables S4A–F and S4I; Fig. S6), with complete sensitivity tables in the package (Supplement S7). The supported correspondence connects feature-defined basin positions with complete profiles of modeled occupancy across repeated transitions in more than one basin solution. Because these solutions aggregate the same source-linked event-states while preserving their narrative adjacencies, variation in correspondence reveals the dependence of the measured relationship on basin construction. The annotation system and preserved mappings thus support both the comparison of representational and sequential organization and the evaluation of how that comparison changes with aggregation.

| **Basin solution** | **Basin pairs** | **Feature–routing r** | **Nominal p** | **Representational–successor r** | **Nominal p** | **All-space Holm p** |
|---|---|---|---|---|---|---|
| 4-D ten-basin (main) | 45 | 0.334 | 0.038* | 0.427 | 0.080 | 0.203 |
| 4-D nine-basin | 36 | 0.400 | 0.016* | 0.415 | 0.110 | 0.203 |
| 4-D six-basin | 15 | 0.798 | 0.016* | 0.832 | 0.005** | 0.044* |
| 2-D nine-basin | 36 | 0.263 | 0.204 | 0.455 | 0.035* | 0.198 |
| 2-D eight-basin | 28 | 0.596 | 0.028* | 0.684 | 0.001** | 0.013* |

Table 3. Feature–routing and representational–successor correspondence across five basin solutions. Feature–routing correlations compare cosine similarities of basin-state feature profiles (mechanism presence, primary context, and event-level meta-theme) with routing-profile similarities. Representational–successor correlations compare 4-D centroid distances with complete successor-profile distances at $\gamma = 0.90$. Nominal p values are unadjusted. All-space Holm p adjusts across five solutions using each solution's largest p across 4-D, 10-D, and 13-D tests (Table S4H). Asterisks use unrounded p values: * $p < 0.05$; ** $p < 0.01$. Basin pairs share states; solutions reuse the same observations.

## 4 Discussion

The present study developed a hippocampal-informed annotation system for constructing source-linked event-states from autobiographical narrative and a quantitative framework for examining their representational and sequential organization. Applied to interviews with five participants, the system produced 256 event-states with categorical feature descriptions and preserved narrative order. Consecutive event-states were closer in the 4-D macro-geometry than expected under shuffled order, and the pooled mean displacement field in Dim0 × Dim3 also differed from its shuffled expectation. In the main 4-D ten-basin solution, similar basin-state feature profiles corresponded with similar one-step routing profiles, while primary context, mechanism presence, and event-level meta-theme differentiated full-system pathway engagement. Pairwise basin-centroid distances were positively associated with corresponding Euclidean distances between complete successor profiles across all five basin solutions, with exploratory support in at least two solutions across 4-D, 10-D, and 13-D geometries. These findings demonstrate how the constructed event representations support comparisons between feature-based organization and narrative progression, including modeled occupancy across repeated transitions.

The annotation system supplies the analytical states required for these comparisons. Event-segmentation and narrative-comprehension accounts explain how experience is partitioned and represented through changes in contextual and situational information (Zacks et al. 2007; Zwaan et al. 1995). Applying these principles to autobiographical transcripts requires operational decisions about which changes establish a new event and which properties characterize that event. The present system expresses those decisions through an event-structure schema and explicit boundary rules, complemented by thematic and affective annotations. Schema-constrained LLM procedures apply these specifications to narrative language, while excerpt identities connect the resulting assignments to their supporting passages. The methodological contribution is the construction of source-linked event-states whose feature representations can be compared with their preserved sequential relationships.

Event-state identity remains distinct from feature similarity. A specific anchor could change while broader categorical descriptions remained stable; at 40 within-segment boundaries, it was the only tracked feature to change. The resulting event-states preserved differences in referential focus that would have been lost if event identity had been defined only by the categorical variables used in MCA. Two event-states can consequently occupy similar positions in the geometry while referring to distinct narrated occurrences with different positions in the sequence. Preserving that distinction allows the framework to ask whether similar events have similar sequential relationships without first treating them as the same occurrence.

Hippocampal models provide a theoretical basis for this construction through their emphasis on contextual binding and the relational organization of experience (Behrens et al. 2018; Bellmund et al. 2018; Eichenbaum 2017; Ranganath 2010). Within the annotation system, these principles guide the selection of text-observable properties and the rules used to organize them. Thematic and affective annotations add information about what an event concerns and its associated valence. The source mappings make the theoretical assumptions inspectable. A boundary or feature assignment can be traced to the passage supporting it, and a revised assignment can be followed through the resulting event representation and its downstream mappings. The hippocampal basis of the schema provides hypotheses about relevant organization that can be evaluated against independent behavioral or neural evidence.

Table 2 illustrates this relationship within a narrative of creative practice. Negotiating an exhibition's resources and describing the exhibited works were represented as distinct event-states within the same meta-theme. The boundary rules distinguish the events, while the shared meta-theme retains their broader thematic relationship. Source mappings allow the event boundaries and feature assignments to be examined against the underlying narrative. The framework provides a method for AI-assisted interpretation of cultural materials in which the distinctions used for quantitative analysis remain connected to their narrative evidence.

The 4-D macro-geometry supplied a compact representation of broad feature organization. Its four macro-clusters showed comparatively strong separation and assignment stability. Macro-clusters differed in feature variables omitted from their corresponding fits, providing evidence of feature relationships beyond those used to construct each solution. Four dimensions were the smallest tested representation supporting four macro-clusters that met the event-count and participant-support criteria. The 10-D participant-robust geometry and 13-D full retained geometry retained additional distinctions for reference comparisons. This distinction is consistent with accounts in which changes in dimensionality preserve different aspects of a representation (Kerrén et al. 2025). Macro-clusters characterized broad feature organization. Basin-states were fitted separately to pool nearby event-states for transition estimation.

The event-level findings connect this feature organization to the progression of narrated recall. Consecutive event-states were 10.7% closer in the 4-D macro-geometry than expected under endpoint-fixed within-participant narrative-order shuffling, indicating local continuity in the represented features. The pooled mean displacement field in Dim0 × Dim3 also differed from its shuffled expectation. These findings are compatible with accounts linking contextual representations to retrieval order and temporal structure to representational organization (Howard and Kahana 2002; Polyn et al. 2009; Schapiro et al. 2013). The present comparison establishes continuity relative to reordered versions of the same narratives. Because shuffling also removes topic persistence and other sequence structure, it does not identify the process producing that continuity. The field result similarly concerns the pooled pattern rather than the significance of individual arrows or a particular trajectory shape.

Basin-state analyses extended the comparison from proximity between consecutive occurrences to relationships among destination distributions. Each event-state contributes at most one observed outgoing transition; pooling nearby event-states provides repeated observations from which a routing profile can be estimated. In the 4-D ten-basin solution, cosine similarities between basin-state feature profiles combining mechanism presence, primary context, and event-level meta-theme were positively associated with cosine similarities between routing profiles. Feature–routing associations were positive across all five basin solutions, although statistical support varied with feature-profile definition and basin construction. The relationship connects basin-state feature composition with the distribution of outgoing transition probabilities across destination basin-states. It gives event-state features a role in describing sequential organization beyond their use in constructing the representational geometry.

Feature-conditioned pathway-engagement comparisons further locate the relationship between represented properties and pathway use. A feature condition can be associated with a different transition pattern because its event-states begin in different basin-states, because destinations differ from a shared source basin-state, or through both processes. Full-system comparisons capture the combined pattern, whereas supported-source comparisons address destination differences at shared sources. Primary context and mechanism presence differentiated full-system pathway engagement across all five basin solutions, with event-level meta-theme supported in the three 4-D solutions. The more consistent evidence concerns which portions of the transition system were used. Supported-source comparisons were nonsignificant in the 4-D ten-basin solution. This distinction gives an operational form to questions about contextual engagement of functional networks raised in prior conceptual work (Sariñana 2025), while identifying denser observations at shared sources as a requirement for stronger routing claims.

Representational–successor correspondence extends the comparison to occupancy relationships implied by repeated transitions. At $\gamma = 0.90$, a complete successor profile includes current-state identity and expected discounted occupancy across subsequent basin-state transitions (Dayan 1993; Gershman et al. 2012). Representational–successor correspondence compared pairwise Euclidean basin-centroid distances with the corresponding Euclidean distances between complete successor profiles. Positive associations indicate that basin-states closer in representational geometry tended to have more similar complete successor profiles. Holm-adjusted tests identified the 4-D six-basin and 2-D eight-basin solutions as supported across 4-D, 10-D, and 13-D centroid-distance geometries. The 4-D ten-basin estimate was positive but nonsignificant. The result provides a naturalistic setting for examining relationships between representational organization and transition-derived structure considered in predictive-map accounts (Stachenfeld, Botvinick, and Gershman 2017). The interpretation concerns complete successor profiles. Profile-definition controls did not establish a unique contribution from accumulated multistep structure beyond immediate transitions and current-state identity.

The variation across basin solutions makes aggregation part of the model requiring evaluation. All five solutions retained the same event-state identities and narrative adjacencies, but changed which occurrences were pooled. Such changes alter basin-state feature profiles and destination distributions, including which event-to-event adjacencies become self-transitions. The 4-D ten-basin solution was selected before routing and successor comparisons using assignment-stability and outgoing-support criteria. Corrected successor support in the 4-D six-basin and 2-D eight-basin solutions addresses the subsequent question of correspondence across retained aggregations. Differences in statistical significance alone do not establish differences between correlations. Preserved mappings nevertheless make the consequences of aggregation inspectable over a fixed set of source-linked observations. The 4-D and 2-D prefixes identify basin fitting in Dim0–Dim3 and Dim0 × Dim3, respectively. Measuring basin-centroid distances in alternative geometries while retaining basin membership separately examines dependence on distance representation.

These source mappings also provide a route from constructed narrative representations to tests of neural organization. Overt autobiographical recall during fMRI can be transcribed and aligned with the BOLD time series (Gilmore, Quach, Kalinowski, González-Araya, et al. 2021; Gilmore, Quach, Kalinowski, Gotts, et al. 2021). Time-aligned source excerpts could locate event-state boundaries and their associated features within a continuously recorded signal. Studies of stable neural event patterns and their organization across cortical regions provide a basis for these comparisons (Baldassano et al. 2017; Geerligs et al. 2022, 2021). Evidence of internally generated context boundaries during recall extends this basis to narrated experience (Lee and Chen 2022). The annotation system would add explicit models of relations among the recalled events, allowing neural analyses to address their organization as well as boundary timing.

Within the same participants, one hypothesis is that event-states with more similar feature representations evoke more similar neural activity patterns. A complementary question is whether one-step transition structure or complete successor profiles explain neural relationships beyond those captured by the feature geometry. Evidence for hippocampal and medial prefrontal contributions to event segmentation and integration motivates comparing these models across regions (Liu et al. 2022). Successor-like representations observed after sequence learning provide a further basis for testing transition-derived occupancy relationships (Ekman, Kusch, and De Lange 2023; Garvert, Dolan, and Behrens 2017). Comparisons with routing profiles and identity-removed successor profiles would help determine which components account for any neural correspondence. Such experiments would evaluate the constructed states as models of recalled experience, with their source passages specifying the narrative evidence associated with each neural epoch.

The annotation architecture could also accommodate schemas informed by other neural systems. An amygdala-informed module could characterize affective significance and target–outcome relations (Janak and Tye 2015; Jin et al. 2015). Schemas informed by accumbens or prefrontal models could describe motivated action or goal maintenance and control demands, respectively (Friedman and Robbins 2022; Klawonn and Malenka 2018). These modules could organize the same passages at different resolutions. Affective significance might change within a contextually stable event, whereas a maintained goal might span several event-states. Alignment through common excerpts would preserve those differences and allow their relationships to be examined. The relevant test would be whether an additional schema explains narrative or neural organization beyond the existing representation.

Several limitations define the next steps in evaluating the framework. Event boundaries and feature assignments are theory-guided, model-generated annotations without independent human coding. The manual audit verified source correspondence for the sampled excerpts. Construct validity requires separate assessment of whether the assigned boundaries and features capture the intended properties. Comparisons among human and model annotations could distinguish disagreements about event boundaries from disagreements about features, then examine how each affects the downstream relationships. Numerical reproducibility currently begins with frozen event-level annotations; the public templates support prospective reimplementation rather than regeneration of identical historical annotations. Evaluating annotation reproducibility and downstream numerical reproducibility requires separate checks.

The proof-of-concept sample contained five participants, and the 251 narrative transitions were pooled across their interviews. Sparse conditioned observations limit participant-specific estimates and comparisons from shared sources. Global condition-label shuffles also do not preserve participant-specific condition frequencies. The five basin solutions reuse the same narratives, and focused corrections apply to the stated comparison families within a broader exploratory analysis. Held-out transition checks estimate routing probabilities from other participants while retaining full-sample MCA coordinates and basin memberships, so they do not evaluate generalization of the complete pipeline to new participants. Finally, narrated adjacency records progression during semi-structured recall, which may be influenced by elicitation and discourse organization. Larger datasets with denser within-participant sequences would support more direct evaluation of feature-dependent routing and the portability of the constructed representations.

Principles derived from hippocampal function can guide the identification and organization of information expressed in natural language into constructed cognitive state spaces. In this framework, hippocampal-informed features provide a relational basis for defining event-states, while explicit mappings connect those event-states to basin-states and macro-clusters. These mappings allow representational and sequential relationships to be examined as the same narrative observations are organized at different resolutions. The broader implication is that models of memory organization can inform both the construction of analytical states and the connections among levels of analysis. Making these transformations explicit brings state construction into model testing, allowing different accounts of remembered experience to be compared against the same source-linked observations.

## 5 Statements

### Acknowledgements

The author thanks Daniel Graziano, Luis R. Sandoval, Caroline A. Runyan, and Thomas J. McHugh for their thoughtful discussion, comments, and feedback on the development of the manuscript.

### Preregistration

This work was not preregistered.

### Funding

The broader interdisciplinary project from which this study developed received support from the Somerville Arts Council and the Malden Cultural Council. A grant from the Council for the Arts at MIT supported the purchase of photographic film used for some photographs in the broader project. No dedicated funding was provided for the statistical analyses or preparation of this manuscript.

### Conflicts of interest

The author declares that they have no competing interests.

### Data Availability

The public reproducibility package (version 1.0.0) is archived on Zenodo at https://doi.org/10.5281/zenodo.23050872 and is available on GitHub at https://github.com/joshua-sarinana/narrative-state-spaces. It includes a privacy-reviewed, de-identified event-level dataset containing the 256 event-states used for the deterministic quantitative analyses, its data dictionary, the annotation codebook, and model-independent annotation-constraint prompt templates. Package contents and reproduction procedures are described in Supplement S7.

Numerical reproduction begins with the frozen event-level annotations. The templates support prospective annotation reimplementation using current models; they are not historical production prompts and do not guarantee identical annotations across model versions. The manual excerpt-traceability audit procedure and confidence calculation are described in Supplement S1, and participant-level audit results are reported in Table S1D. Recordings, complete transcripts, the full excerpt and specific-anchor datasets, contextual narrative descriptions, original identifiers, model interactions, and underlying excerpt-level audit records are withheld from public distribution.

### Code Availability

The Python code for the quantitative analyses and computational figure panels, together with the de-identified public event-level input, is available in the public reproducibility package (version 1.0.0), archived at https://doi.org/10.5281/zenodo.23050872. The code repository is available at https://github.com/joshua-sarinana/narrative-state-spaces. The reproducibility package includes run configurations, dependency specifications, reference outputs, consolidated extended tables in Markdown format, code that generates detailed analysis tables in CSV format, and verification procedures (Supplement S7). Numerical reproduction begins with the frozen event-level annotations and does not rerun the upstream LLM annotation procedures.

# Supplementary materials

## Event Memory and Narrative Dynamics

Supplementary Tables S1–S4 are supplied in the companion table document. They report the analytical inputs, construction diagnostics, transition analyses, and successor analyses for the current dataset of 256 event-states. Extended methods below explain how source-linked event construction supports the comparisons of representational organization and narrative sequence. Supplementary Figures S1–S6 and their legends are supplied separately.

Table S1A identifies the 11 active MCA variables and their implementation fields. Complete feature definitions, permitted categories, and event-construction rules are provided in the public annotation codebook within the public reproducibility package (Supplement S7). Descriptive variable names are used throughout this text; implementation fields are retained where needed to identify dataset columns.

## S1 Annotation and event structure system

The annotation pipeline first applied thematic analysis to interviewee speech. The large language model (LLM) identified meaning-bearing passages and retained them as ordered verbatim excerpt units, each intended to be self-contained and contain at least one complete propositional unit. Sentiment was assigned at the excerpt level. The final excerpt text, identifiers, and mappings form the fixed source record for downstream analyses. A separate hippocampal-informed event-structure analysis then used these excerpts as evidence, grouping consecutive excerpts into bounded events and assigning the resulting event-level features. One event-state could contain one or more excerpts. Thematic and sentiment assignments did not determine event boundaries or event-structure feature values. The model-independent annotation-constraint prompt templates for prospective thematic annotation and event construction are described in Supplement S7.

Interviews were divided into processing segments of approximately 10 pages so that each LLM interaction operated on a manageable amount of source material. Event boundaries and features were derived from the interviewee speech within each processing segment. Processing-segment labels recorded source provenance and did not enter the feature schema or boundary rules. The resulting event sequences were subsequently connected across segments within each participant's narrative.

Anchors were described at two levels. The specific anchor identified the fine-grained referential focus organizing an event, while the anchor category classified that focus using predefined categories. Specific anchors were event-specific and were not persistent entity identifiers across the interview. A new event began when the primary mechanism, primary context, specific anchor, or anchor category changed, following the definitions and assignment rules in the public annotation codebook (Supplement S7). Specific-anchor changes could distinguish events when the broader categorical features remained stable.

Representational domain described the dominant mode through which the event content was represented, using the categories visual, auditory, tactile, or semantic. A change in representational domain could contribute to an event boundary only when the source text supported a change in the dominant mode and the new mode dominated at least one complete excerpt; isolated sensory words or secondary representational details were insufficient. Orientation, perspective, reference frame, landmark class, secondary mechanism, and secondary context characterized an event-state but did not independently trigger a boundary.

Specific-anchor text and representational domain were retained in the annotation records for event construction and boundary checks but were excluded from the active MCA variables. Orientation was included in MCA. The 11 active variables comprised nine categorical event-structure variables together with event-level meta-theme and sentiment aggregated from the mapped excerpts (Table S1A).

The LLM models used are reported in main-text Methods 2.2. Annotation required fixed identifiers, one permitted value per categorical field, and excerpt-level provenance. Event-structure features were assigned from interviewee speech using the hippocampal-informed criteria. Thematic and sentiment assignments and downstream analytical results were excluded from this stage to prevent them from influencing event construction or its feature assignments. Audit checks covered required fields, allowed values, identifier consistency, complete excerpt-to-event assignment, and correspondence with source excerpts. These checks assessed procedural consistency and source traceability; they did not independently validate the event ontology or feature assignments.

The procedure is termed hippocampal-informed event-structure analysis because its feature system was informed by principles and network functionality from episodic-memory and navigation models. These theory-guided analytical features organize narrated event content. They do not measure hippocampal activity or participant-specific neural states.

The MCA-ready input contains 20 columns, comprising 11 active variables and nine metadata fields. The metadata preserve participant, event, and processing-segment identity, narrative sequence, and annotation metadata and counts for joining and auditing records; they do not enter MCA. Together, participant_id_norm and hippo_id_norm identify each event-state consistently across the linked datasets. The complete field inventory is provided in the public reproducibility package (Supplement S7). Table S1B reports participant-level interview duration and analytical-unit counts; Table S1C reports descriptive within-segment boundary changes.

Event construction produced 256 event-states within 21 processing segments, yielding 235 within-segment boundaries across five participants. Table S1C summarizes changes in the five tracked boundary features. Counts overlap because multiple features could change at the same boundary. Specific anchor changed at 234 boundaries and was the only tracked feature to change at 40 of the 235 boundaries. At the remaining boundary without an anchor change, primary mechanism, primary context, and representational domain changed. These counts describe how the boundary-rule features changed at established boundaries; they do not quantify each criterion's independent contribution. The 16 links added between processing segments connected separately constructed event sequences and were outside this within-segment boundary assessment.

## Manual excerpt-traceability audit

The author manually assessed a participant-stratified random sample of 150 excerpts from the final 442-excerpt dataset. Sample sizes were allocated proportionally across participants, with integer rounding (Table S1D). Unique worksheet row numbers were drawn separately within each participant's integrated thematic dataset using the CalculatorSoup random number generator (https://www.calculatorsoup.com/calculators/statistics/random-number-generator.php). The selected row numbers and corresponding excerpt IDs were retained. Each selected excerpt was compared against the original interview transcript. Failure was defined as inability to locate the excerpt in its source transcript or a discrepancy beyond trivial formatting differences.

The confidence calculation treated the five participant datasets as fixed finite populations, with simple random sampling without replacement within each participant. For participant h, let N_h denote the number of excerpts, n_h the number sampled, and D_h a hypothetical number of failures. The probability of observing zero failures across all five samples was:

$$P_0(D_1, \dots, D_5) = \prod_{h=1}^{5} \frac{\binom{N_h - D_h}{n_h}}{\binom{N_h}{n_h}}.$$

For each hypothetical dataset-wide failure count D, this probability was maximized over all feasible allocations whose failure counts sum to D. This maximization allowed failure proportions to differ across participants. The conservative one-sided 95% upper confidence limit was the largest integer D for which the maximized probability of observing zero failures was at least 0.05.

All 150 sampled excerpts passed. The maximized zero-failure probability was 0.05518 for seven hypothetical failures and 0.03623 for eight. The upper confidence limit was therefore seven failures among 442 excerpts, or 1.58%. This bound concerns excerpt-level source correspondence within the analyzed dataset; event-boundary and feature-assignment validity require separate assessment.

## S2 Data construction and integration

The analyzed corpus comprised five completed full interviews conducted over six sessions (Table S1B); introductory interviews were excluded (main-text Methods 2.1). Linking the segment-level outputs preserved their narrative order, excerpt identifiers, event definitions, feature assignments, and processing-segment provenance. Three linked components connected the segment-level outputs through ordered thematic excerpt records, event-level records, and an explicit excerpt-to-event mapping. The mapping assigned each of 442 excerpt units to one of 256 event-states across 21 processing segments (Table S1B).

### Meta-theme construction and mapping

The 62 distinct superordinate-theme labels collected across participants were each assigned to one of 12 broader meta-themes. Grouping used the meanings of the labels without their frequencies, participant counts, or downstream analytical results. Each meta-theme was defined broadly enough to accommodate additional superordinate themes.

The mapping was applied to excerpt records by a deterministic Python lookup. Join keys were obtained from superordinate-theme labels (superordinate_theme) by trimming leading and trailing whitespace, collapsing internal whitespace, and converting text to lowercase. A many-to-one left join associated each excerpt record with its mapped meta-theme. Subsequent codebook-based checks assessed mapping agreement, allowed categorical values, and uniqueness of participant–excerpt keys.

The thematic component of the public annotation codebook was compiled after mapping and documented the resulting classifications. It defines the 12 meta-themes and records their associated superordinate-theme labels (Supplement S7). This mapping provided shared meta-theme categories across participants while preserving the original excerpt identifiers.

## Excerpt-to-event join and aggregation

Thematic records were linked to the excerpt-to-event mapping through participant and excerpt identifiers; participant and mapped event identifiers then linked these records to the event-level feature table. Participant and processing-segment provenance were retained. Each excerpt was assigned to one event-state; an event-state could receive one or more consecutive excerpts. The join associated each event with the thematic labels of its mapped excerpts while retaining the event-structure features assigned directly to that event. Meta-theme and sentiment were then aggregated separately within each event. The most frequent value was retained, with ties resolved lexicographically by the values' string representations. The resulting event-level table (256 events) contained one row per event-state, including event-level meta-theme, event-level sentiment, and the number of mapped excerpts. Across all events, the excerpt counts summed to 442.

## Sequence reconstruction and analysis input

Within each participant, next-event links (hippo_next_global) connected event records across processing segments in narrative order. Together with the preceding-event links (hippo_prev_global), these fields preserve five participant-specific sequences and 251 consecutive event-state adjacencies, including 16 links between processing segments (Table S1B; analysis workbook field inventory in the public reproducibility package, Supplement S7). Processing-segment identity is provenance metadata rather than an analytical state variable.

The MCA-ready workbook contains 256 rows and 20 columns in Sheet1. In addition to the 11 active MCA variables, its nine metadata columns are participant_id_norm, hippo_id_norm, hippo_id, uid, hippo_weight_tag, hippo_prev_global, hippo_next_global, n_theme_assignments_event, and n_excerpts_event (Table S1A; analysis workbook field inventory in the public reproducibility package, Supplement S7). The original analysis workbook and the de-identified public input have different schemas, as detailed in Supplement S7. The public input additionally retains representational-domain labels and a privacy-preserving anchor-change flag for boundary summaries; specific-anchor text remains in the protected annotation records.

Mappings from excerpts to event-states and from event-states to macro-clusters or basin-states preserve source traceability across aggregation levels. Every event-state remains linked to its constituent excerpts, and every subsequent macro-cluster or basin-state assignment remains linked to the event-state from which it was derived. Alternative basin solutions reorganize the same event-states and original narrative adjacencies rather than creating a new underlying dataset. The sequence and provenance fields are identified in the analysis workbook field inventory in the public reproducibility package (Supplement S7); the analytical-unit counts are reported in Table S1B.

## Interview scaffold and adaptive probing

The autobiographical interview used a semi-structured scaffold rather than a fixed question schedule. Core domains included developmental and geographic background, community membership and belonging, spatial navigation, creative practice, professional activity, digital-media use, and interactions with social and physical environments. The scaffold also contained optional probes concerning embodied action, adaptation to uncertainty, and the meanings participants assigned to particular experiences.

The interviewer adapted wording, order, and follow-up questions to each participant's background and responses. Participant-specific notes and examples informed some probes, and the written questions were not administered verbatim as a standardized schedule. The scaffold defined the intended domains of elicitation.

## S3 Representational geometry

Multiple correspondence analysis (MCA) used 11 active categorical variables, comprising nine event-structure variables and event-level meta-theme and sentiment aggregated from excerpt annotations (Table S1A). The specific anchor and representational domain contributed to event construction but were not active MCA variables. The full fit produced 54 dimensions. The 13-D full retained geometry is the full-sample statistically retained reference, and the 10-D participant-robust geometry is the reference dimensionality supported after each participant exclusion. The 4-D geometry was selected for the principal analyses. These geometries use the first 13, 10, and four dimensions, respectively, of the same full-sample MCA solution. With zero-based axis labels, they span Dim0–Dim12, Dim0–Dim9, and Dim0–Dim3. Dimensionality retention and selection of a macro-cluster representation answer separate questions.

Parallel analysis was implemented by independently permuting each active variable within participant, preserving participant event counts and participant-by-variable category margins while disrupting within-event cross-variable associations. Retention was the longest leading sequence whose observed eigenvalues exceeded the rank-matched 95th percentiles of 5,000 null spectra. The full sample retained Dim0–Dim12; Dim13 failed the criterion. Participant-exclusion MCA and parallel-analysis refits retained at least 10 dimensions in every fold, establishing the 10-D geometry as the participant-robust reference. Full-sample and participant-exclusion retention counts are reported in Table S2A. Figure S1A,D shows full-sample and participant-exclusion retention, respectively.

Table S2B reports the raw and positive Benzécri-corrected inertia captured by each named geometry (Benzécri 1979) (Fig. S1C). Raw percentages are relative to total raw inertia; corrected percentages are relative to total positive Benzécri-corrected inertia. Full-geometry fidelity compares the named geometries with the complete 54-dimensional solution (Fig. S1B; Table S2B). For full-geometry fidelity, neighborhood comparisons explicitly excluded each event itself and used a fixed order of participant and event identifiers to resolve numerical distance ties within an absolute tolerance of $10^{-10}$. Spearman correlations used average ranks for tied distances. The complete tie convention is documented in the public reproducibility package (Supplement S7).

Projection selection evaluated all axis pairs within each nested full-sample representation from four through 13 dimensions. Mean neighborhood trustworthiness across neighborhood sizes of 5, 10, and 20 event-states was the primary criterion, followed by Spearman and Pearson correspondence between projected and reference-space distances. Dim0 × Dim3 was selected for the 4-D geometry. Dim0 × Dim5 was selected for both the 10-D and 13-D reference geometries. Table S2B reports the selected projection for each named reference geometry; complete candidate diagnostics are generated by the public reproducibility workflow (Supplement S7). The common Dim0 × Dim3 projection used in the manuscript follows the selection for the selected 4-D geometry.

Projection checks that refitted MCA after participant exclusion selected Dim0 × Dim3 in three of five folds under a common 10-D reference and two of five folds under fold-specific retained dimensionality.

Because MCA was re-estimated after each participant omission, its axes could change. These checks assess participant influence on projection selection. The current basin solutions were fitted in either four dimensions or the Dim0 × Dim3 plane. Within each solution, memberships were held fixed when centroid distances were compared across the 4-D, 10-D, and 13-D geometries.

The macro-organization comparison evaluated two through 13 dimensions and two through six clusters without using transition or successor outcomes. This comparison assessed geometric separation, event-resampling stability using 20 bootstrap samples, participant support, MCA-refit participant-exclusion stability, and held-out construct alignment. k-means used 10 initializations and seed 42 for the full fits. Eligibility required a minimum cluster fraction of 0.05 and at least two participants per cluster. The four-cluster solution in the 4-D geometry was selected as the compact four-cluster representation for the principal macro-scale comparisons. Complete grid diagnostics are generated by the public reproducibility workflow (Supplement S7); Table S2C reports selected comparisons with lower-dimensional, fewer-cluster alternatives and with higher-dimensional four-cluster models. Macro-model selection was separate from dimensionality retention and basin-resolution selection.

Held-out construct analysis assigned the active variables to six predefined sets of related feature variables, comprising context (primary and secondary context), mechanism (primary and secondary mechanism), anchor and landmark (anchor category and landmark class), representational organization (reference frame, orientation, and perspective), event-level meta-theme, and event-level sentiment. Each complete set was removed before MCA and k-means were refitted. Held-out categorical variables were represented by centered one-hot indicator blocks, each scaled to equal total sum of squares within its set. Multivariate pseudo-F measured separation of the held-out profiles across the resulting macro-clusters (Anderson 2001); $R^2$ measured the explained variance. The complete held-out profile was shuffled jointly within participant using 10,000 permutations while macro-cluster labels remained fixed. Null-corrected $R^2$ was observed $R^2$ minus its null mean. The manuscript reports Benjamini–Hochberg (BH) q values (Benjamini and Hochberg 1995) across the six feature sets within the four-cluster solution in the 4-D geometry; the separate grid-wide adjustment remains in the source data. These tests excluded feature variables from model construction and assessed their correspondence with the resulting macro-clusters.

Four dimensions were the smallest tested representation in which a four-cluster model satisfied both support criteria. Four-cluster models in two and three dimensions each contained a single-event cluster. The selected 4-D model contained 133, 59, 17, and 47 event-states, with at least four participants in every cluster. Its additional descriptive distinction included a separate virtual/digital cluster (Table S2D). Among eligible four-cluster models in the focused grid, the 4-D model had the greatest silhouette and bootstrap assignment agreement (Table S2C).

The matched four-cluster comparisons clarify the choice of 4-D over the higher-dimensional references. Silhouette was 0.349 in 4-D, 0.165 in 10-D, and 0.122 in 13-D. Corresponding bootstrap adjusted Rand indices (ARI) were 0.735, 0.474, and 0.470, and mean participant-exclusion ARI values were 0.720, 0.592, and 0.503. Mean held-out null-corrected $R^2$ values were similar at 0.054, 0.055, and 0.051. These descriptive diagnostics favored a compact, more clearly separated and reproducible four-cluster representation; no significance test compared these diagnostic values.

The neighboring 5-D four-cluster solution also had lower geometric separation and participant-exclusion assignment agreement than the selected 4-D solution (Table S2C). Extended macro-model comparisons are provided in the public reproducibility package (Supplement S7).

The 10-D and 13-D geometries retained more overall representational information. Their distance correlations with the complete 54-dimensional geometry were 0.864 and 0.909, compared with 0.738 for 4-D (Table S2B). Higher-dimensional models could also support coarser organizations. The 13-D two-cluster model, for example, had a mean participant-exclusion ARI of 0.908 (Table S2C). The preference for 4-D concerns the intended four-cluster macro scale and does not establish that dimensions beyond four are noise or that 4-D is optimal for every analysis. The 10-D and 13-D geometries were retained as reference distance spaces. Basin solutions were fitted only in 4-D or the selected 2-D projection, so these results do not compare basin fits in 10-D or 13-D.

Feature enrichment characterized C1 as social context integration/belonging, C2 as goal-directed creative/prospective organization, C3 as virtual-digital mediation, and C4 as physical-spatial place/path organization. These descriptions summarize fitted macro-cluster composition and were not supplied to k-means. Five of the six held-out feature sets were significantly aligned after correction; event-level sentiment was not (Table S2D, accompanying note).

## S4 Basin state construction and resolution selection

Basin-state construction used size-constrained clustering while retaining the original MCA scaling. The 4-D basin solutions were fitted in Dim0–Dim3, and the 2-D basin solutions were fitted in Dim0 × Dim3. Identical normalized profiles across the 11 active features remained indivisible and were weighted by their numbers of event-states. For fixed centroids, binary mixed-integer optimization assigned profiles within lower and upper event-count bounds; weighted-centroid updates then iterated the solution. Among starts, the lowest geometric sum of squared errors satisfying the outgoing-support rule was retained. If no start satisfied support, the geometric best was flagged unsupported. The outgoing criterion influenced solution selection, although the fixed-centroid assignment objective used geometry alone. Successor or routing outcomes did not enter selection. The construction-settings table in the public reproducibility package reports the complete settings (Supplement S7).

Basin centroids are the mean coordinates of their assigned event-states. The 4-D basin-centroid distances supply the main representational–successor analysis described in S6 for every solution, including the solutions fitted in Dim0 × Dim3. The 10-D participant-robust geometry and 13-D full retained geometry provide higher-dimensional references. Dim0 × Dim3 centroids supply the common projection and matched two-dimensional Procrustes comparison. The fitting space, centroid-distance geometry, and visualization space are distinct.

Participant-stratified event omissions assessed assignment stability with fixed MCA coordinates and with MCA refitted to the retained events, following the use of perturbations to assess partition reproducibility. Adjusted Rand index (ARI) summarized agreement with the full-sample partition, and the minimum matched-state Jaccard summarized the least reproducible basin-state in each trial (Hennig 2007). The pilot used ten two-event and ten 13-event omissions with 20 starts per fit. Validation used the remaining 40 two-event and 90 13-event omissions, with 30 starts per fit; a separate omission of the two event-states added during the chronological correction used 100 starts. Reference labels did not initialize omission fits. Pilot and validation sample identifiers were disjoint, but their overlapping event subsets and shared participants do not provide independent validation (Table S2G).

## Operating-resolution selection

The screen requested 2,496 configurations across two fitting spaces, five through 20 basin-states, lower event-count bounds of eight, nine, or ten, and every integer upper bound from 25 through 50. Each feasible configuration used ten independent initializations, supplemented by available solutions from the preceding upper-bound setting. Arithmetic-infeasible and equality-only configurations were recorded separately. The screen produced 2,082 successfully fitted configurations, of which 2,030 satisfied their occupancy bounds and retained at least eight outgoing observations per basin-state. These are configuration counts; different bound settings could produce identical event-state partitions. The staged selection procedure is summarized in Figure 3A; complete construction settings are reported in the public reproducibility package (Supplement S7).

Geometric screening compared eligible configurations within each fitting space and requested basin count using mean fitting-space silhouette (Rousseeuw 1987), the lowest basin-mean silhouette, minimum outgoing support, and the fraction of event-states assigned beyond their nearest centroid. Higher values were preferred for the first three measures and lower values for the last. Nondominated configurations were prioritized. For these configurations, no other configuration in the comparison set was at least as favorable on every measure and more favorable on at least one. The mean of the four percentile ranks then ordered candidates, giving each measure equal weight. One representative per fitting space and basin count was retained, supplemented where necessary to represent each fitting-space and lower-bound combination. Duplicate partitions were removed, leaving 31 pilot candidates.

The lowest basin-mean silhouette was calculated by averaging event-state silhouettes within each basin-state and taking the smallest basin average in the configuration. It identifies the least-separated basin-state relative to the others in that solution; no positive silhouette cutoff was required for eligibility. Figure S3A displays this measure against the fraction of variation in the common 4-D coordinates accounted for by the basin centroids. The horizontal measure describes geometric coverage and was not separately included among the screening criteria. Silhouettes were calculated in each solution's fitting space, so the 2-D and 4-D values describe different geometries. The plotted eligible configurations therefore show geometric tradeoffs rather than a ranking determined by proximity to the upper-right corner.

Each pilot candidate was refined using 100 independent full-sample initializations, with its retained screen partition included as an additional candidate. Pilot omission analyses then evaluated reproducibility using the schemes described above. Ranking used the 13-event refits and combined mean ARI, the 10th percentile of ARI, mean minimum matched-state Jaccard, and the fraction of refits retaining adequate outgoing support. The two-event omissions supplied a smaller-perturbation check and contributed to completion eligibility. Within each fitting space, nondominated stability profiles were prioritized and the four percentile ranks were averaged. Candidates required an eligible full-sample fit and at least 90% completion in each pilot omission scheme. Partitions with ARI $\geq$ 0.95 against an already retained representative were skipped. This procedure advanced five 4-D and three 2-D candidates to the reserved omission analyses.

Final eligibility required at least 95% fit completion and retention of at least eight outgoing observations per basin-state in at least 80% of trials in each refitted omission scheme. Eligible candidates were ranked within fitting space using the same stability measures, now calculated from the 90 reserved 13-event omissions. The diversity rule was reapplied, retaining up to three 4-D and two 2-D solutions. Remaining ranking ties were resolved by geometric sum of squared errors, count-bound settings, and configuration identifier. Five was therefore the specified maximum number of retained solutions rather than a number inferred from the data. The separate omission of the two event-states added during chronological correction did not contribute to ranking.

The retained set comprised the selected 4-D ten-basin solution with event-count bounds of 9–47, the 4-D nine-basin and six-basin solutions with bounds of 8–49 and 8–50, and the 2-D nine-basin and eight-basin solutions with bounds of 8–50 and 8–39, respectively (Table S2E). Geometric superiority on the two axes of Figure S3A did not guarantee retention. For example, the 2-D fourteen-basin candidate with bounds of 10–44 exceeded both retained 2-D solutions on these plotted measures and passed the final support requirement. It nevertheless ranked lower under the stability criteria. Its mean refit ARI was 0.509, compared with 0.575 and 0.521 for the retained nine-basin and eight-basin solutions, and its minimum matched-state Jaccard was substantially lower on average. It was consequently not retained within the two-solution allocation for that fitting space.

The selected 4-D ten-basin solution had realized event counts of 9–47 and minimum outgoing support of nine. Its mean ARI under the 90 reserved refitted 13-event omissions was 0.733, with outgoing support retained in 87/90 trials. Local separation remained uneven. B3 contained nine event-states from three participants and had a mean fitting-space silhouette of −0.238. Table S2G reports assignment agreement and outgoing-support retention under the reserved omission analyses for all five selected solutions. The staged procedure favored reproducibility among screened candidates without establishing that the retained partitions were uniquely optimal. Its thresholds, ranking rules, and solution allocations were analyst-specified exploratory choices. The shortlist was fixed before downstream routing and successor results; this procedure was not preregistered.

Table S2F reports modal primary context and primary mechanism, event counts, participant coverage, outgoing support, and self-transition probabilities for the ten basin-states in the selected solution. These feature summaries are descriptive and do not define membership. Basin numbers are local to each solution; B1 in one solution is not assumed to identify the same set of event-states in another. Assignment stability and minimum outgoing support do not establish precise estimates for every transition probability.

### Post-selection basin-solution sensitivity

Basin-solution sensitivity preserved the same 256 event-states and 251 within-participant adjacencies while changing the event-to-basin assignment. Every event-state received one assignment within each solution, and each transition system was recomputed by mapping the same original adjacencies through that assignment. The partitions were not required to form a nested tree. They provide dependent sensitivity views of the same dataset, not additional participant samples.

## S5 Transition dynamics and permutation tests

Transitions followed the next-event links within each participant, yielding 251 event-state adjacencies, comprising 235 within processing segments and 16 between segments. They represent adjacency in narrated recall rather than temporal adjacency in lived experience. Event-state order, MCA coordinates, macro-cluster membership, and basin assignments were held fixed except for the order manipulations specified below.

## Consecutive event-state distances

To test whether adjacent event-states were locally similar in the constructed feature space, Euclidean distances were calculated from the unstandardized Dim0–Dim3 coordinates for all 251 observed adjacencies. The primary statistic was the pooled mean distance, weighting each adjacency equally. Longer narratives consequently contributed more observations. No macro-cluster or basin assignment entered this test.

The null uniformly permuted interior event order independently within each participant while fixing the first and last event-states. It preserved participant event sets and source and destination margins, while removing interior sequence structure, including persistence across topics and processing-segment boundaries. The exact expected total distance for a participant was the sum of distances from the first event-state to every interior event-state, from every interior event-state to the last event-state, and between all ordered pairs of distinct interior event-states, divided by the number of interior event-states. Each permitted adjacency has that reciprocal inclusion probability. Dividing the expected total by the participant's number of adjacencies gave its expected mean; the pooled expectation weighted these participant means by adjacency counts.

The two-sided statistic was the absolute difference between the observed mean and this exact null expectation. It was compared with the corresponding absolute differences from 1,000,000 randomizations, using seed 20260926 and independent PCG64 child streams for the five participants. The p-value was $(b + 1)/(1{,}000{,}000 + 1)$, where b counted null statistics at least as large as observed, with a numerical tie tolerance of $10^{-12}$ scaled by the larger of 1 and the observed absolute difference. The plus-one correction follows Phipson and Smyth (2010). This was one exploratory primary test without a within-analysis multiplicity adjustment. Participant summaries were descriptive.

Consecutive event-states were 10.7% closer than the exact shuffled-order expectation in the constructed 4-D geometry ($p = 1 \times 10^{-6}$, rounded at the plus-one Monte Carlo resolution limit). No randomized statistic was at least as extreme as the observed absolute departure, and all five participant means were below their corresponding expectations (Table S3D; Fig. 2B). This comparison is conditional on the constructed event boundaries and geometry. Euclidean step distances are unchanged by reversing a whole sequence, so the test does not establish direction-specific organization.

## Projected mean displacement field

The Dim0 × Dim3 field summarized the mean one-step displacement from source to destination event-states. Source counts and summed displacements were accumulated on a 25 × 25 grid spanning the event coordinates with 5% padding. Gaussian smoothing used a standard deviation of 0.75 grid cells and nearest-value boundary handling. Smoothed displacement sums were divided by smoothed source counts, retaining cells with smoothed counts of at least 0.05. Context- and anchor-category-conditioned fields used the same 25 × 25 grid, smoothing, and support threshold. All-event density backgrounds used a 25 × 25 grid and a Gaussian standard deviation of 1.0 grid cells. These conditioned fields remained descriptive (Fig. S2).

A global test compared the pooled observed field with the exact expected field under endpoint-fixed interior-order shuffling. This reference accounts for the inward tendency generated by random ordering in a bounded coordinate distribution. The expected successor of the first event-state was the mean of all interior coordinates. For each interior event-state, the expectation averaged the other interior coordinates together with the last event-state's coordinates. Expected displacements were processed using the same grid, smoothing, and source-count division. The statistic was the mean squared vector difference between observed and expected fields, weighted by smoothed source counts normalized over the fixed retained cells. Grid cells were not treated as independent observations.

The field test and five macro-to-next-basin tests described below used the same 1,000,000 endpoint-fixed within-participant randomizations, generated with PCG64 seed 20260926. Upper-tail p-values used the plus-one formula and numerical tie tolerance of $10^{-12}$ scaled by the larger of 1 and the observed statistic. Holm correction (Holm 1979) covered this six-test family. The global field departure remained supported after correction (Table S3E; Fig. 2C–D). The inference concerns the pooled field and supplies no separate tests of individual cells, arrows, or conditioned fields.

## Basin transition probabilities and macro-scale comparisons

Mapping each event adjacency to the selected 4-D ten-basin solution produced the 10 × 10 count matrix C. Dividing each row by its sum produced the empirical probability matrix P without smoothing and with self-transitions retained. A basin-state's routing profile was its row of P, containing its one-step probabilities to every destination basin-state, including itself. The same count and normalization procedure was applied to all five basin solutions. Any zero-outgoing row in a sensitivity analysis remained zero. Counts were pooled across five participant sequences because individual sequences were too sparse for high-precision participant-specific matrices. Basin support is reported in Table S2F; the numerical source files contain the full count and probability matrices.

At the macro scale, counts were aggregated under the fixed four-cluster memberships in the 4-D geometry and row-normalized with within-cluster transitions included in the denominators. The distance comparison retained the 12 observed nonzero directed between-cluster pairs. Euclidean distances between those cluster centroids were calculated separately in the 4-D, 10-D, and 13-D geometries, holding cluster assignments fixed. Two-sided inference enumerated all 24 centroid-label permutations. All three distance–transition comparisons were nonsignificant (Table S3A).

## Macro-cluster membership and subsequent basin-state membership

The temporal relationship between broad and finer aggregations was assessed using conditional mutual information between source macro-cluster and destination basin-state, conditional on participant. For each of the five basin solutions, a participant-specific count matrix contained four source macro-cluster rows and one column per destination basin-state. Mutual information used natural logarithms and was averaged across participants with weights proportional to their numbers of adjacencies. Within-label persistence was included. The statistic was compared with the shared endpoint-fixed randomizations used for the field test; its null mean provided the reference for finite-sample positive bias.

Conditional mutual information exceeded the shuffled-order reference in the three 4-D basin solutions after Holm correction across the six tests (Table S3E). These tests assess departure from randomized interior order within the observed narratives. They do not isolate particular macro-to-basin links or separate the contribution of macro-cluster identity from current basin-state identity.

## Incremental held-out prediction of the next basin-state

A separate descriptive analysis asked whether current basin-state identity improved prediction of the next basin-state beyond current macro-cluster identity. One participant at a time was excluded when estimating transition probabilities, and predictions were scored on that participant's adjacencies. MCA coordinates and all assignments remained fixed from the full dataset, so the held-out evaluation applied to transition estimation rather than the entire representation-building procedure.

A global destination distribution was estimated from training counts with a total prior mass of 1 distributed uniformly over destination basin-states. The macro-cluster predictor combined its destination counts with this global distribution using a fixed shrinkage mass of 10. The macro-cluster-plus-basin-state predictor combined counts for each source combination with the corresponding macro-cluster predictor, also using a shrinkage mass of 10. An unseen source combination consequently fell back to the macro-cluster predictor. No smoothing parameter was tuned. The mean log probability ratio of the combined predictor to the macro-cluster predictor, in nats per held-out adjacency, quantified the gain using the logarithmic scoring rule (Gneiting and Raftery 2007).

Adding current basin-state identity gave negative pooled prediction gains in all five solutions, and no participant showed a positive gain. Table S3F reports the estimates and training support. Support was summarized using a threshold of at least eight training adjacencies from at least three training participants per source combination. These thresholds were descriptive heuristics. No significance test was applied because unrestricted order shuffling would also destroy the macro-level transition structure needed for an incremental comparison.

## Feature–routing correspondence and feature-conditioned pathway engagement

The feature profiles used in the primary feature–routing comparison concatenated counts for mechanism presence, primary context, and event-level meta-theme. Mechanism presence combined the primary and secondary mechanism variables, ignored whether a mechanism was assigned as primary or secondary, and counted each distinct mechanism once per event. Feature-profile similarity was cosine similarity between basin feature-count vectors; routing-profile similarity was cosine similarity between corresponding rows of P. For each solution, correlations used the unique unordered pairs of distinct basin-states. A common row-and-column permutation of the routing-similarity matrix preserved its internal dependence while the feature-similarity matrix remained fixed. Two-sided tests used 10,000 permutations, seed 42, and $(b + 1)/(10{,}000 + 1)$, where b counted absolute correlations at least as large as observed. Comparisons were repeated across all five solutions using all 11 active MCA variables, retaining primary and secondary mechanism as separate variables in this alternative feature definition. Individual p-values were nominal and unadjusted across basin solutions and feature definitions (Fig. 3C-D; Fig. S4; Table S3B).

The feature–routing association in the selected solution remained positive after each participant omission, with statistical support varying across omissions (Table S3B, accompanying note). These checks held the fitted representation and assignments fixed while rebuilding pooled profiles from the retained participants. They assess participant influence within the existing dataset and do not validate the complete procedure on independent participants.

Conditioned routing grouped frozen transitions by the source event's primary context, mechanism presence, or event-level meta-theme. Feature conditions required at least 10 source transitions, retaining four contexts, six mechanisms, and 11 meta-themes. Condition matrices were row-normalized. The full-system statistic was mean pairwise cosine similarity between flattened condition matrices. The supported-source statistic compared destination distributions only from source basins with at least three outgoing transitions under both conditions being compared.

Categorical condition labels were shuffled globally across source transitions 10,000 times with seed 42. Primary-secondary mechanism pairs were shuffled jointly before overlapping mechanism-presence groups were rebuilt. Condition matrices and shared-source support masks were recomputed within each shuffle. Lower-tail tests asked whether observed similarity was smaller than under the null. BH correction was applied across the three feature variables separately for full-system and supported-source tests within each basin solution. Global shuffling does not preserve participant-specific condition frequencies.

Primary context, mechanism presence, and event-level meta-theme differentiated full-system pathway engagement in the selected 4-D ten-basin solution. Supported-source comparisons did not establish differences in destination distributions from shared adequately sampled source basin-states in this solution. The supported-source mechanism result in the 2-D nine-basin comparison depended on participant composition. Table S3C reports all five solutions and summarizes support in its note; detailed condition-support summaries are provided in the public reproducibility package (Supplement S7).

## S6 Successor structure and robustness

The standard successor representation was $M = (I - \gamma P)^{-1} = I + \gamma P + \gamma^2 P^2 + \cdots$, evaluated by solving $(I - \gamma P)M = I$ directly (Dayan 1993; Stachenfeld, Botvinick, and Gershman 2017). The primary analysis used $\gamma = 0.90$; discount checks used 0.70, 0.80, and 0.95. Each row is a complete successor profile with one entry per basin-state in that solution. I assigns weight 1 to the current basin-state before a transition occurs, and the remaining terms encode expected discounted subsequent occupancy. Complete successor profiles include current-state identity; the identity-removed comparison removes this contribution.

The main representational–successor analysis calculated the Pearson correlation between Euclidean distances among 4-D basin centroids and Euclidean distances among complete successor profiles. The selected 4-D ten-basin solution supplied 45 unique basin pairs. The same 4-D distance analysis was applied to all five solutions, including the solutions fitted in two dimensions. The 10-D and 13-D geometries supplied reference centroid distances while basin memberships and successor profiles remained fixed. Distances within each pair of matrices were matched by basin identity; the two spaces did not need equal dimensionality. Basin pairs share states and are not independent participant observations. Two-sided common-label matrix tests (Mantel 1967) applied the same row-and-column permutation to the successor-distance matrix while geometric distances remained fixed, using 10,000 draws with replacement and seed 42. The p-value was $(b + 1)/(10{,}000 + 1)$, where b counted absolute null correlations at least as large as the observed absolute correlation. These individual p-values are unadjusted across solutions, geometries, discount factors, and sensitivity checks; the focused across-solution assessments below use separate combination procedures. Complete-profile estimates are reported in Table S4A; discount-factor checks are provided in the public reproducibility package (Supplement S7).

Principal component analysis (PCA) of the column-centered successor matrix supplied a two-dimensional visualization. Matched two-dimensional Procrustes analysis (Gower 1966) aligned Dim0 × Dim3 basin centroids with Successor PC1 × PC2 through centering, normalization, and orthogonal rotation/reflection. Alignment used 10,000 common-label permutations with a lower-tail disparity test. Table S4B reports the explained variance, disparity, and permutation results for the selected and 2-D nine-basin solutions (Fig. 5A; Fig. S5A–C); all five solutions are reported in the public reproducibility package (Supplement S7). This projected-map comparison is separate from the complete-profile distance analysis in Figure 5C and Figure S5D.

The two-component successor display retained 36.1% of successor-profile variance in the selected solution. Distances in this display and distances between complete successor profiles define different comparisons; the reduced display does not substitute for the complete-profile analysis. The projection-restricted comparison is reported in the accompanying note to Table S4B.

At $\gamma = 0.90$, 4-D basin-centroid distances were positively associated with complete successor-profile distances across all five basin solutions, with nominal statistical support varying by solution. Estimates for all three centroid-distance geometries are reported in Table S4A; the complete discount-factor sweep is provided in the public reproducibility package (Supplement S7). Different significance statuses do not establish differences between correlations, and the strongest association did not determine the selected solution.

Participant exclusion rebuilt P and M from the retained participant sequences. One comparison held full-sample basin centroids fixed to isolate transition-system influence; a complementary comparison recomputed centroids from retained events. Both retained the original MCA coordinates and basin memberships. Table S4C summarizes omission ranges and counts of nominally significant tests; complete omission-specific estimates are provided in the public reproducibility package (Supplement S7). These overlapping pooled-system checks assess participant influence with the representation and basin memberships held fixed.

Basin-visit sensitivity collapsed each maximal consecutive run within the same basin-state to one visit and rebuilt P and M from the resulting visit sequence. The underlying event identities, features, and memberships were retained. One discounted step then represented a transition between consecutive basin visits, changing the effective horizon in event units. All five basin-visit comparisons were nominally significant, with descriptively larger correlations in the selected and 2-D nine-basin solutions (Table S4D; Fig. S6A–B). Correlations did not increase uniformly across solutions. Separate tests of these correlations do not establish a significant increase or isolate a causal effect of within-basin persistence.

Post hoc order controls held event identities, coordinates, features, basin assignments, participant membership, and sequence lengths fixed. Reversal reversed each complete participant sequence. In 10,000 null draws, event order was independently shuffled within participant and P and M were rebuilt. These shuffles allowed the first and last event-states to change, unlike the endpoint-fixed null used for the event-distance, field, and macro-to-next-basin tests in S5. Upper-tail order-shuffle tests compared correspondence in narrated and reversed orders with the shuffled-order null. Figure S6D displays both correlations with the central 95% of individual-solution shuffle-null correlations. This interval is a null reference range rather than a confidence interval. Table S4E reports the estimates and tests; complete order-control results are provided in the public reproducibility package (Supplement S7). These order controls assess a different null hypothesis from the basin-label tests. Their separate p-values do not test a difference between narrated and reversed correlations.

Profile-definition controls compared 4-D basin-centroid distances with distances between rows of P, $I + 0.90P$, standard M, and $M - I$, holding memberships fixed within each of the five basin solutions (Table S4F; Fig. S6C). The P comparison uses Euclidean routing-profile distances and differs from the cosine feature–routing comparison in Figure 3D. Identity alone gives constant distances between distinct basis vectors. These comparisons did not establish a unique contribution from accumulated multistep structure. Their separate p-values do not test differences between correlations.

### Across-solution correspondence and joint narrated-order inference

The across-solution family comprised the five basin solutions in Table S2E, holding complete successor profiles, $\gamma = 0.90$, and event adjacencies fixed. For $m = 5$ ordered component p-values, the Bonferroni partial-conjunction test for at least u non-null basin solutions used $\min[1, (m - u + 1)p_{(u)}]$. This conservative procedure allows arbitrary dependence between valid component tests (Wang and Owen 2019). Cumulative maxima of these p-values as u increased implemented fixed-sequence testing and a lower bound on the number of supported basin solutions. Holm adjustment (Holm 1979) across the five basin solutions separately identified supported solutions (Table S4H).

To require correspondence throughout the named distance spaces, the largest of each basin solution's 4-D, 10-D, and 13-D component p-values supplied its intersection–union test. Partial conjunction was then applied across these five maxima. This requires support in the same basin solutions across all three spaces. The primary and all-space claims are nested; separate distance-space summaries are sensitivity checks. These procedures do not correct across every endpoint or every candidate basin solution considered during model construction.

As a precision check, all 720 basin-label permutations for the six-basin solution and all 40,320 permutations for the eight-basin solution were enumerated in each distance space. Exact p-values used the fraction at least as extreme as the observed absolute correlation, including the identity permutation and numerical ties within $10^{-12}$. Larger solutions retained the existing 10,000-draw Monte Carlo estimates. The rule for exact enumeration depended on basin count. Both versions supported at least two basin solutions in 4-D ($p = 0.022$) and across all three spaces ($p = 0.044$); neither supported at least three (Table S4G). For the six-basin 4-D comparison, 4/720 rounds to $p = 0.006$, whereas its original Monte Carlo estimate rounds to $p = 0.005$. Original individual tables retain the latter estimate; the cross-solution results retain both calculation versions.

Joint narrated-order inference followed the principle of evaluating a fixed cross-specification summary against shared randomized datasets (Simonsohn, Simmons, and Nelson 2020; Winkler et al. 2016). The statistic was fixed as the median Pearson correlation across all five basin solutions before joint results were calculated. Each draw permuted event positions independently within the five participant-specific sequences and applied those same position permutations to every basin solution. Event identities, coordinates, basin memberships, participant occupancy, and sequence lengths were held fixed. P and M were rebuilt for each basin solution. Original adjacency and segment-specific composition were not preserved; the existing sequence convention spanning each participant's processing segments includes 16 between-segment joins. These shared order shuffles were unrestricted within participants, allowing endpoints to change. Table S4I summarizes the geometry-specific joint tests; their null distributions are generated by the public reproducibility workflow (Supplement S7).

Upper-tail joint p-values used (1 + number of null medians at least as large as the observed median)/(10,000 + 1), including numerical ties within $10^{-12}$. The joint 4-D and 10-D tests met the nominal 0.05 criterion, but the requirement for support across all three geometries was not met (maximum joint $p = 0.054$; Table S4I). This order-null assessment differs from the basin-label partial-conjunction test and does not identify an order effect in every basin solution or a unique accumulated multistep contribution.

## S7 Reproducibility and materials

Public numerical reproduction begins with a de-identified, 256-row event-level CSV and its data dictionary. This 19-field public input is distinct from the original 20-column analysis workbook described in Supplement S2, Table S1A, and the analysis workbook field inventory in the public reproducibility package. It retains the 11 active MCA variables, stable replacement identifiers, sequence links within each participant, excerpt counts, representational-domain labels, and a derived anchor-change flag for boundary summaries. Preceding- and next-event links preserve 251 adjacencies, including 16 between-segment joins. Restricting sequence reconstruction to within-segment links would omit those joins and change the transition system.

The public reproducibility package contains the frozen input, public annotation codebook, analysis and computational-figure scripts, run configurations, dependency specifications, reference outputs, and verification procedures. The codebook is provided at documentation/Annotation_Codebook.xlsx. Its HPC_Codebook and Event_Construction sheets define the event-structure categories and boundary rules; Thematic_Domains and Theme_Examples define the meta-themes and their associated superordinate-theme labels. The analysis code covers MCA geometry, macro-cluster comparisons, basin construction and perturbation checks, transition measures, successor analyses, and their controls. Reference outputs are comparison targets for verification rather than inputs to the calculations being checked. Fixed input versions, random seeds, numerical settings, and output checks support numerical reproduction. The workflow starts from annotated event-level records and does not regenerate or independently validate the upstream annotations. The public reproducibility package (version 1.0.0) is archived on Zenodo at https://doi.org/10.5281/zenodo.23050872 and is available on GitHub at https://github.com/joshua-sarinana/narrative-state-spaces.

Tables S1–S4 accompany the manuscript. The public reproducibility package provides 13 consolidated extended tables in documentation/EXTENDED_RESULTS.md. Executing the analysis pipeline generates the detailed analysis tables as CSV files under outputs/. The file documentation/RESULT_TO_CODE.md maps the reported results to their executing code and output locations.

The consolidated extended tables document the original analysis workbook and corpus units, macro-cluster comparisons, basin construction settings, and feature-conditioned routing. They also report successor analyses across discount factors and distance geometries, Procrustes alignment, participant omissions, event-adjacency and basin-visit comparisons, narrative-order controls, and profile-definition controls. An additional table reports exploratory basin recurrence. The public 19-field event-level input is documented separately in data/DATA_DICTIONARY.csv.

The public reproducibility package includes model-independent annotation-constraint prompt templates for prospective annotation, together with the public annotation codebook. The templates specify required inputs, boundary constraints, and output requirements. They document constraints derived from the annotation workflow and do not constitute a verbatim release of the historical production prompts. They are not intended to regenerate identical annotations across model versions. Annotation documentation and fixed numerical inputs serve different reproducibility purposes.

The manual excerpt-traceability audit procedure and confidence calculation are described in Supplement S1, and participant-level audit results are reported in Table S1D. Recordings, complete transcripts, the full excerpt and specific-anchor datasets, contextual narrative descriptions, original identifiers, model interactions, and underlying excerpt-level audit records are withheld from public distribution. The main-text worked example presents selected illustrative passages and anchor descriptions; it does not release the complete source datasets.

## Additional references

References also cited in the main manuscript are listed there. Additional references used in the supplementary methods are provided below.

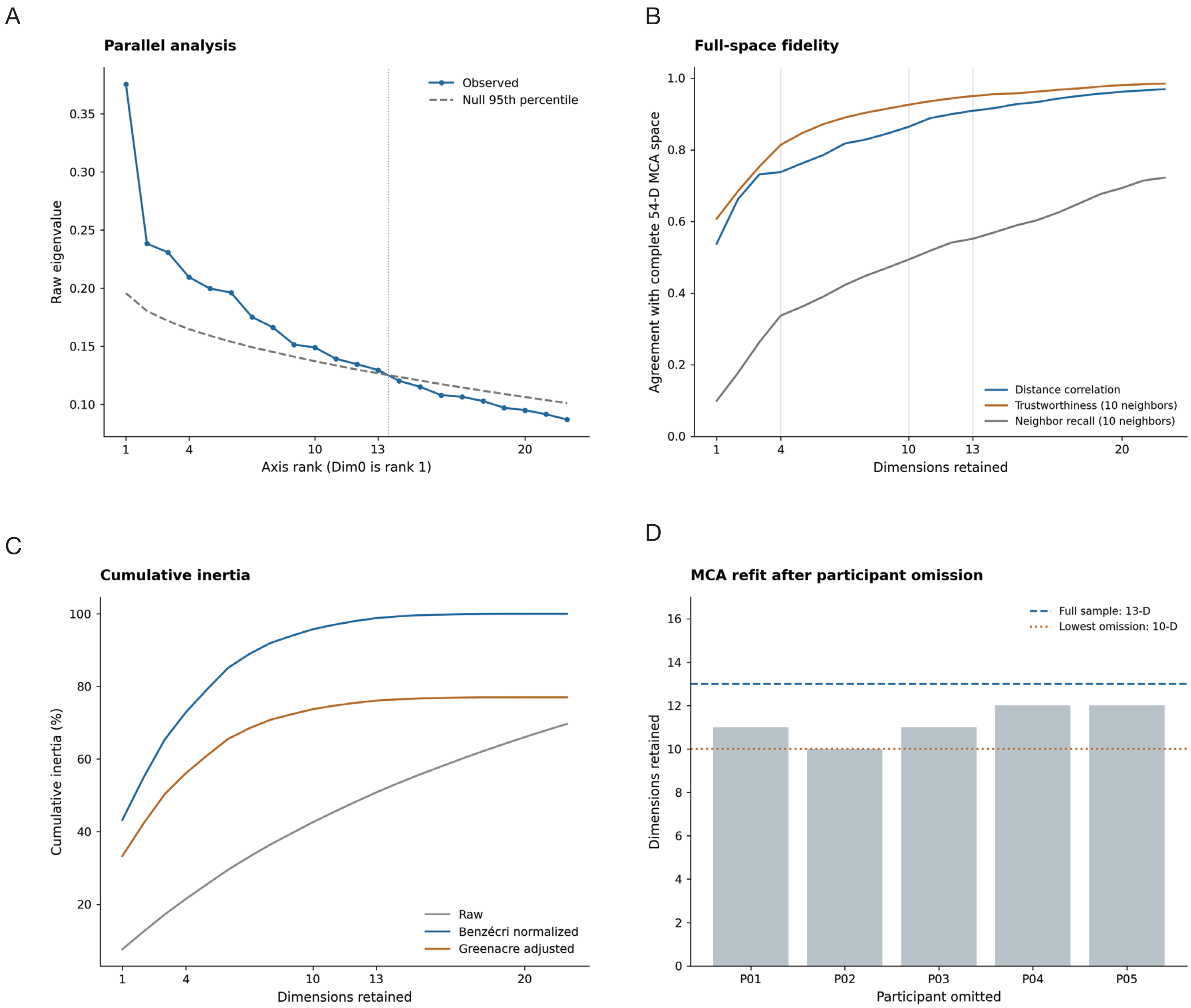

**Figure S1. MCA dimensionality and representational fidelity. (A)** Observed eigenvalues and the rank-matched 95th-percentile parallel-analysis reference. **(B)** Agreement with the complete 54-D MCA solution, measured by pairwise-distance Pearson correlation, neighborhood trustworthiness, and neighbor recall. Neighborhood measures use 10 neighbors; vertical guides mark 4, 10, and 13 dimensions. **(C)** Cumulative inertia under raw, Benzécri-normalized, and Greenacre-adjusted conventions. **(D)** Dimensions retained after omitting each participant and refitting the MCA. Dimensionality retention and selected-geometry summaries are reported in Tables S2A–B.

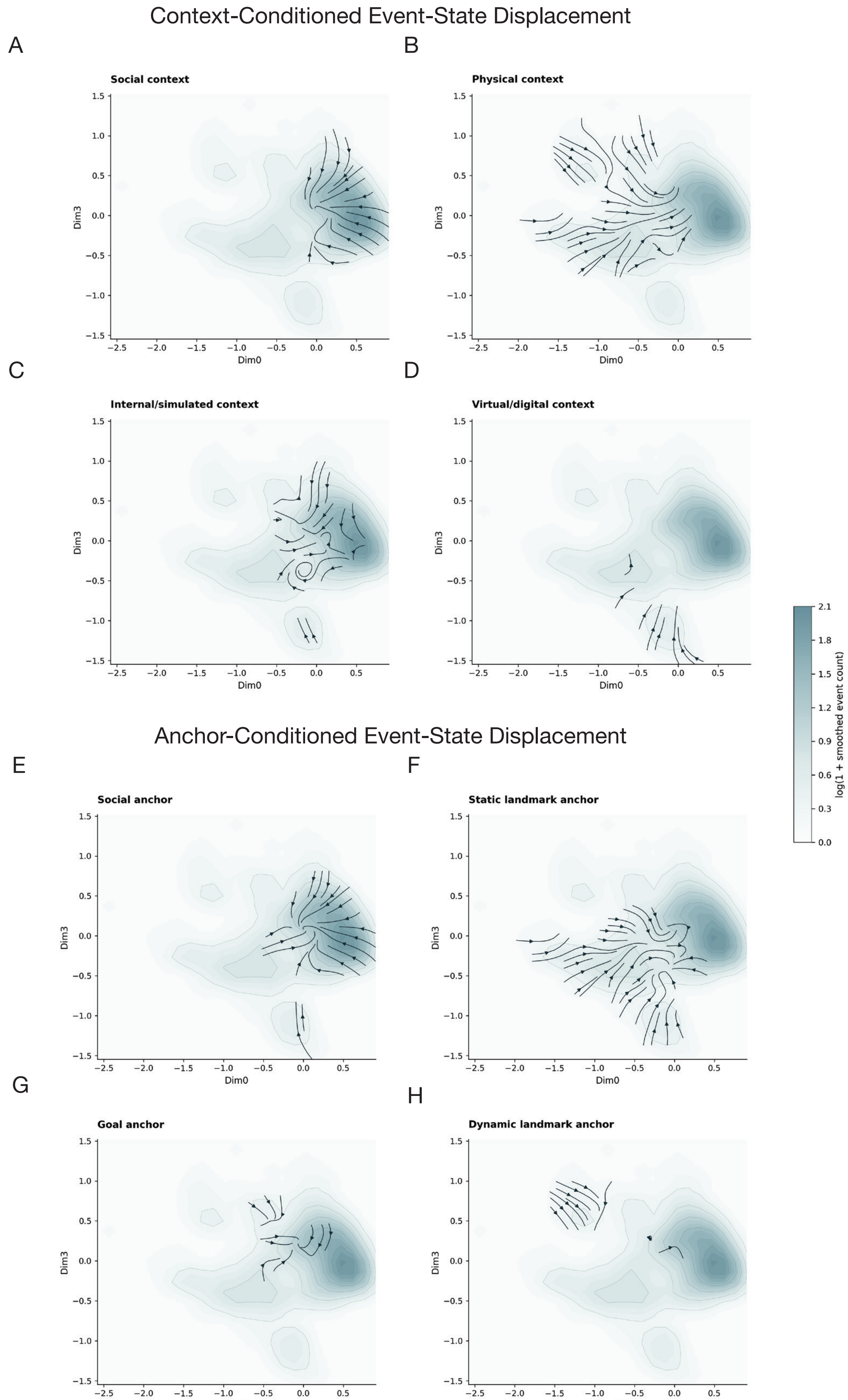

**Figure S2. Context- and anchor-conditioned event-state displacement. (A–D)** show fields conditioned on social, physical, internal/simulated, and virtual/digital source contexts. Source-transition counts are 143, 64, 30, and 14. **(E–H)** show social, static-landmark, goal, and dynamic-landmark source anchors. Source-transition counts are 146, 66, 18, and 15. All panels use the same Dim0 × Dim3 projection, 25 × 25 grid, coordinate limits, and density background representing all 256 event-states. Gaussian smoothing uses 0.75 grid cells for displacement sums and source counts and 1.0 grid cell for density; fields are displayed where smoothed source counts are at least 0.05. Streamlines summarize local mean one-step displacement rather than recorded multistep trajectories. Condition-specific significance was not tested. Pooled observed and exact expected shuffled fields appear in main Figure 2.

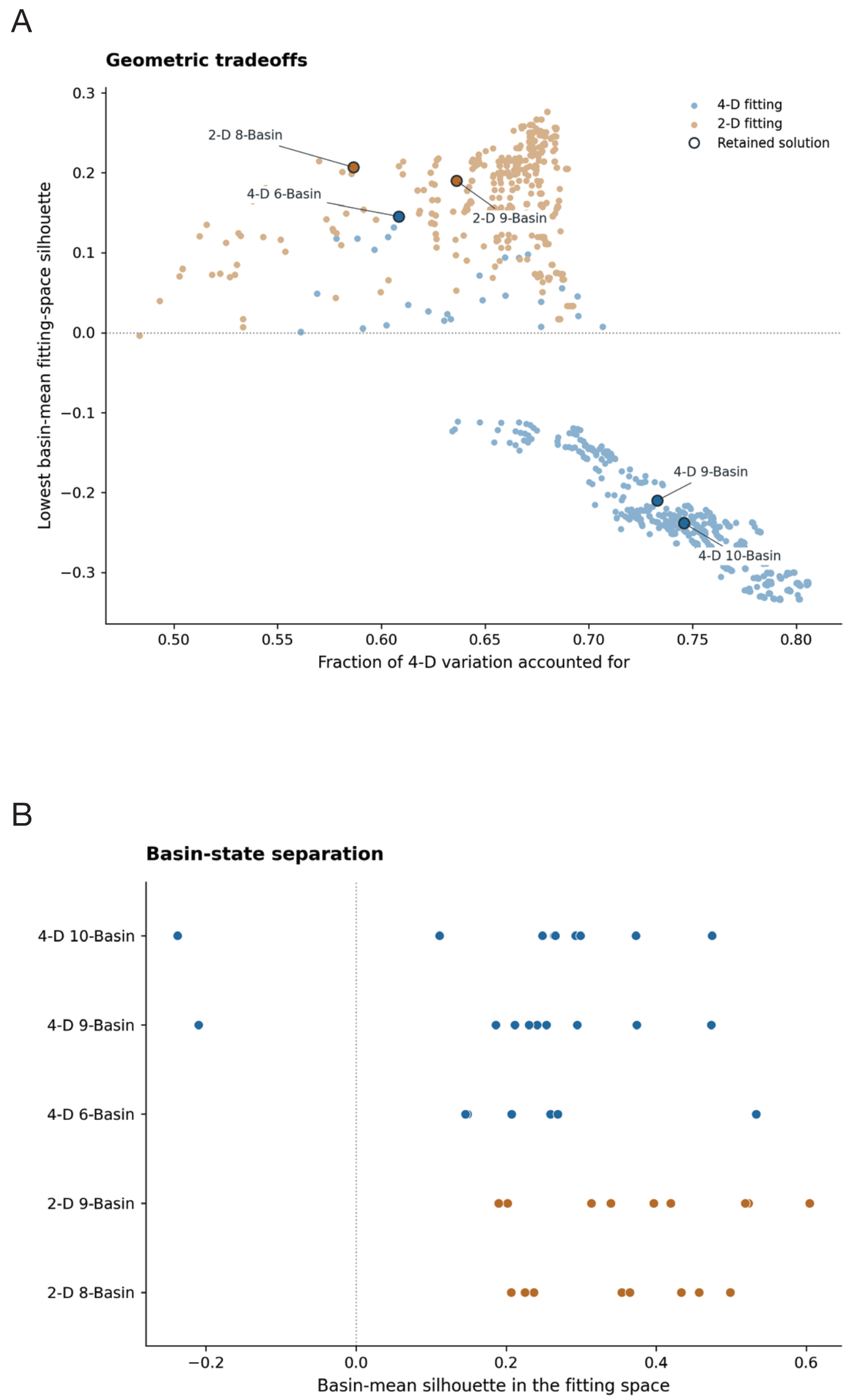

**Figure S3. Geometric tradeoffs and basin-state composition. (A)** The 2,030 eligible construction configurations are shown as equally opaque circles; the five retained solutions are slightly larger outlined circles. The horizontal axis shows the fraction of variation in common 4-D coordinates accounted for by basin centroids. The vertical axis shows the lowest basin-mean silhouette, calculated in each solution's fitting space. It identifies the least-separated basin-state within that configuration; zero is a reference rather than an eligibility cutoff. The 2-D and 4-D silhouettes describe different fitting geometries. Retention also depended on assignment stability, outgoing support and partition diversity. **(B)** Basin-mean silhouettes in each solution's fitting space. Basin labels are local to a solution.

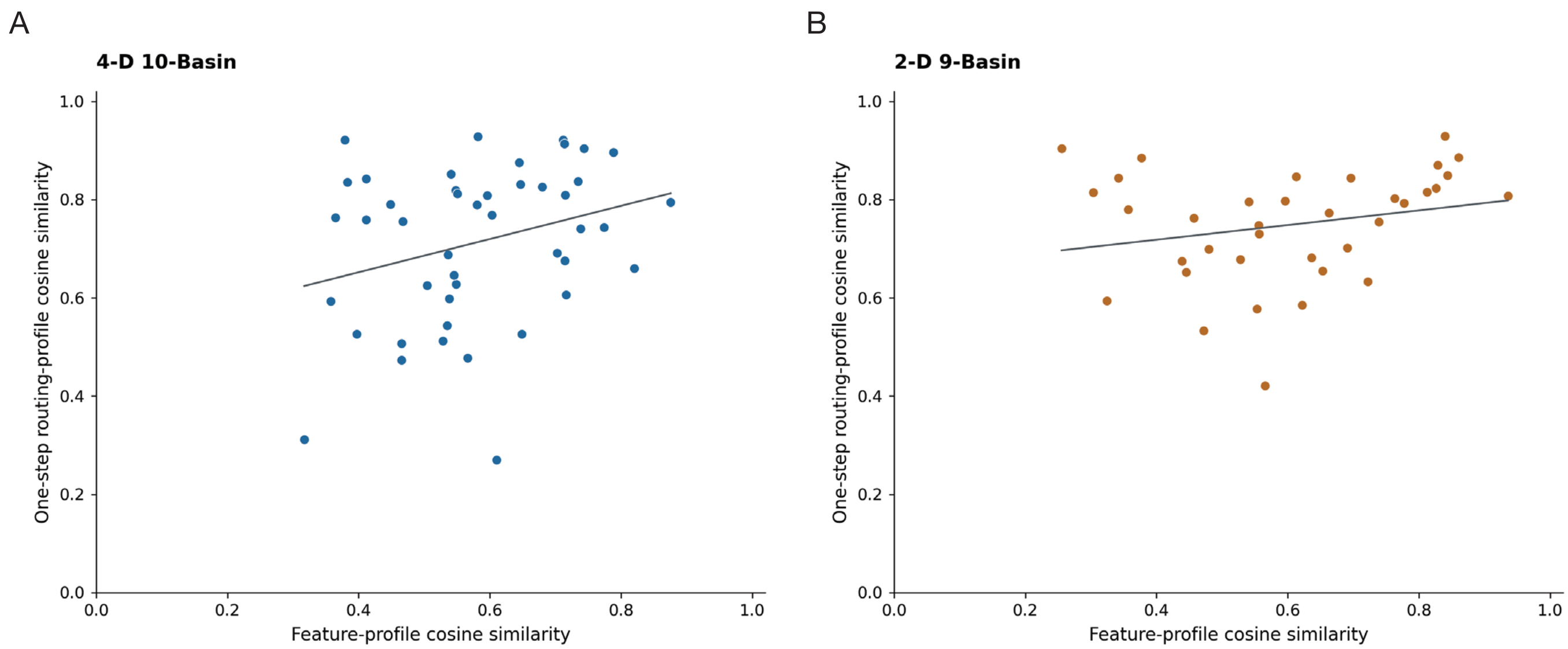

**Figure S4. Feature–routing correspondence using all 11 active MCA variables. Pairwise feature-profile and one-step routing-profile cosine similarities. (A)** 4-D 10-Basin solution ($r = 0.290$, $p = 0.053$) and **(B)** 2-D 9-Basin solution ($r = 0.234$, $p = 0.224$). Each point represents a basin pair, with 45 pairs in A and 36 in B. Pairs share basin-states, and fitted lines are descriptive. These sensitivity profiles include all 11 active MCA variables, retaining primary and secondary mechanism fields separately. Tests use 10,000 two-sided common-label permutations; p-values are nominal and unadjusted across solutions (Table S3B).

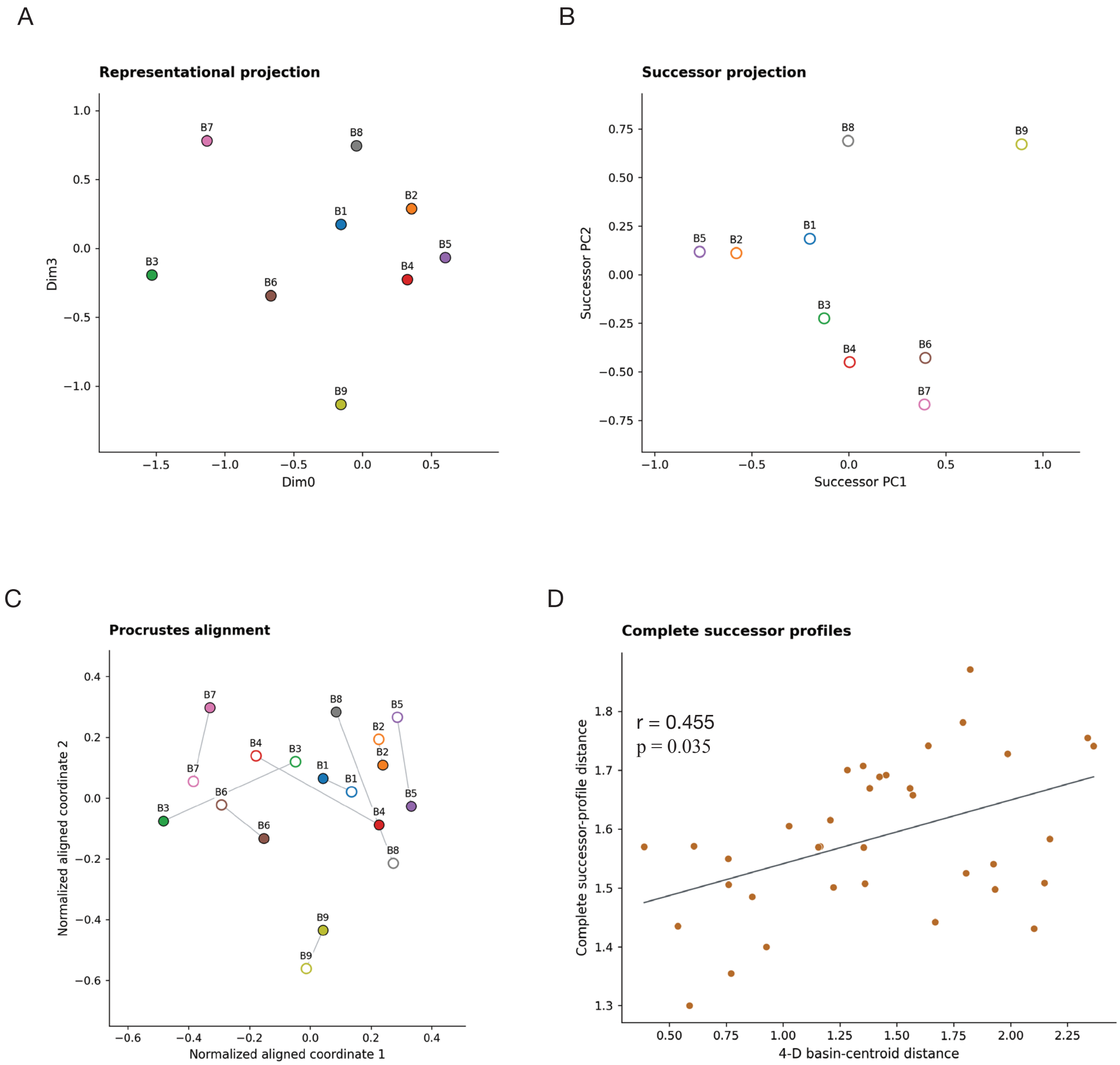

**Figure S5. Representational–successor correspondence in the 2-D 9-Basin solution. (A)** Basin centroids in the Dim0 × Dim3 projection. **(B)** Complete successor profiles displayed in Successor PC1 × PC2 at $\gamma = 0.90$. The two components retain 39.1% of successor-profile variance. Matching colors identify basin-states within this solution. **(C)** Two-dimensional Procrustes alignment: filled circles denote normalized representational centroids, and open circles denote aligned successor positions. Connecting segments show alignment residuals. Procrustes disparity is 0.943, with lower-tail $p = 0.209$. **(D)** Pairwise 4-D basin-centroid distances versus complete successor-profile distances across 36 basin pairs, with $r = 0.455$ and nominal two-sided $p = 0.035$. The fitted line is descriptive. Both tests use 10,000 common-label permutations. Basin fitting uses two dimensions, while D measures centroid distances in four dimensions.

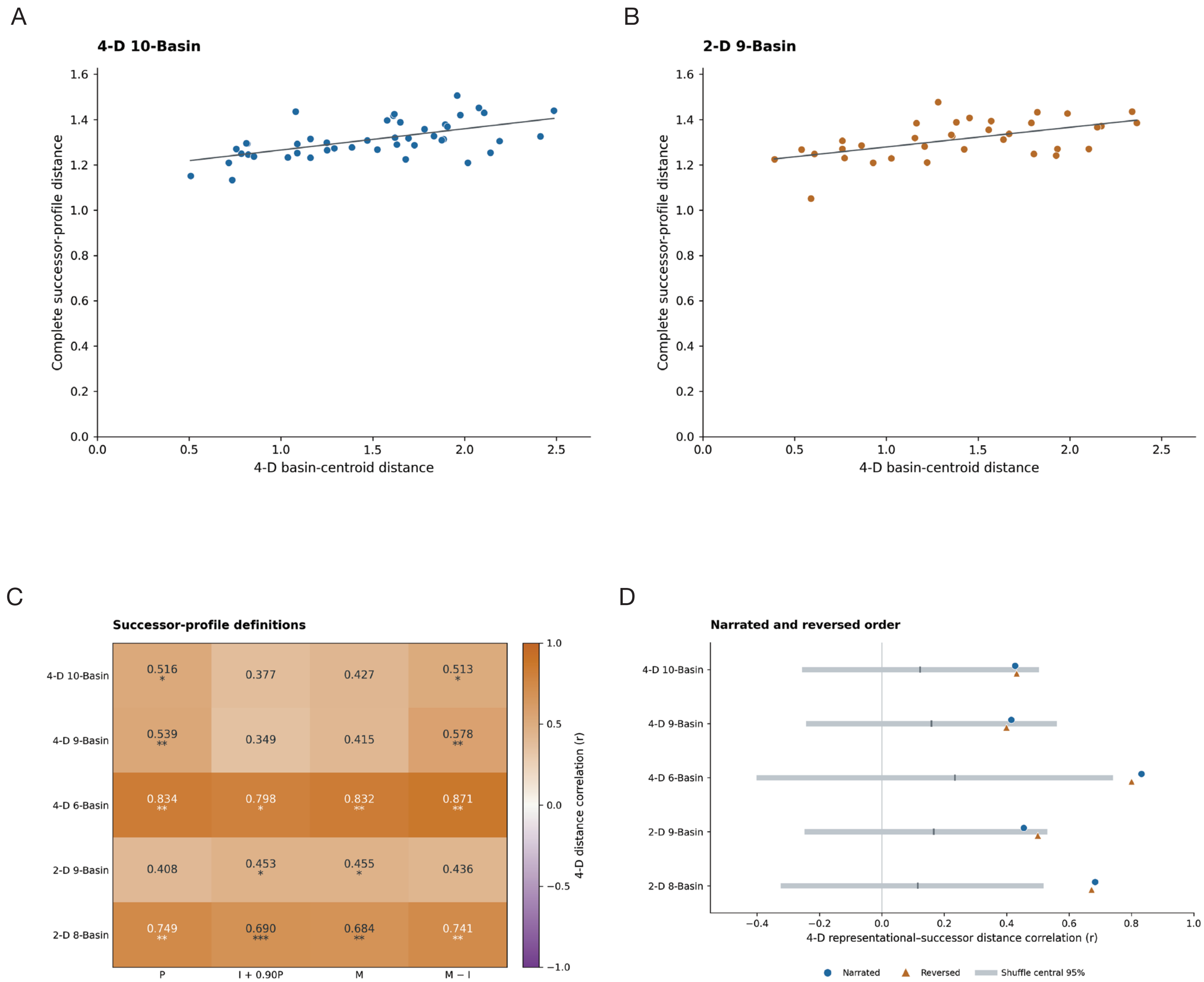

**Figure S6. Transition-unit, profile-definition and narrative-order controls. (A–B)** Complete successor-profile distance correspondence after collapsing consecutive event-states assigned to the same basin-state into single basin visits, shown for 4-D 10-Basin and 2-D 9-Basin, respectively. The comparison changes the transition unit from consecutive event-states to successive basin visits. **(C)** Representational correspondence for one-step routing, identity plus discounted one-step routing, complete successor profiles, and successor profiles with the identity term removed. Color denotes Pearson r. Asterisks denote nominal two-sided common-label permutation p values: $^{*}p < 0.05$, $^{**}p < 0.01$, $^{***}p < 0.001$. **(D)** Correspondence for narrated and reversed order, compared with the central 95% of within-participant shuffled-order results. The gray ranges describe the shuffle distribution. Sequence reversal preserves adjacency while changing direction. These controls do not by themselves establish a unique multistep contribution or forward-direction specificity. Distances use the 4-D geometry and successor-based comparisons use $\gamma = 0.90$.

# Supplementary Tables

Event Memory and Narrative Dynamics: A Large Language Model Framework for Multidimensional State Spaces

Tables S1–S4 report the evidence supporting event construction, representational organization, narrative transitions, and successor profiles for 256 event-states from five participants. The accompanying reproducibility package regenerates detailed diagnostics and complete sensitivity results as CSV tables.

## Table S1 Variables participants and event boundaries

### A Active categorical variables

| Variable | Implementation field | Definition / origin |
|---|---|---|
| Primary mechanism | primary_hipp_mech | Dominant theory-guided mechanism category; event level. |
| Secondary mechanism | secondary_hipp_mech | Additional mechanism category; event level. |
| Primary context | context_type_primary | Dominant context type; event level. |
| Secondary context | context_type_secondary | Additional context type; event level. |
| Anchor category | anchor_type | Coarse category of the event's referential focus. |
| Reference frame | frame | Event-level reference-frame category. |
| Orientation | orient | Event-level orientation category. |
| Perspective | perspective | Event-level perspective category. |
| Landmark class | landmark | Event-level landmark category. |
| Event-level meta-theme | event_meta_superordinate_theme_primary | Cross-participant meta-theme, aggregated from excerpts. |
| Event-level sentiment | event_sentiment_primary | Sentiment category, aggregated from excerpts. |

The 11 variables enter MCA. Identifiers, sequence links, and descriptive fields do not. Complete field definitions are provided in the repository data dictionary and annotation codebook.

# Table S1 Variables participants and event boundaries

## B Participant characteristics and corpus counts

| ID | Sessions | Duration | Excerpts | Event-states | Event-state transitions | Basin-states occupied |
|---|---|---|---|---|---|---|
| P01 | 1 | 1 h 38 min | 52 | 29 | 28 | 10 |
| P02 | 1 | 2 h 20 min | 84 | 56 | 55 | 9 |
| P03 | 2 | 3 h 25 min | 136 | 66 | 65 | 9 |
| P04 | 1 | 1 h 40 min | 59 | 34 | 33 | 9 |
| P05 | 1 | 2 h 22 min | 111 | 71 | 70 | 10 |
| Total | 6 | 11 h 25 min | 442 | 256 | 251 | 10 retained overall |

Five completed full interviews were conducted over six sessions; introductory interviews were excluded. The 442 excerpts map to 256 event-states across 21 processing segments. The 251 participant-global adjacencies comprise 235 within-segment boundaries and 16 between-segment links.

Basin occupancy refers to the main 4-D ten-basin solution. The total reports distinct basin-states, not the sum of participant occupancy counts.

## C Feature changes at within-segment event boundaries

| Feature change | Boundaries, n | Percentage of 235 boundaries |
|---|---|---|
| Specific anchor changed | 234 | 99.6 |
| Primary mechanism changed | 151 | 64.3 |
| Anchor category changed | 115 | 48.9 |
| Primary context changed | 111 | 47.2 |
| Representational domain changed | 53 | 22.6 |
| At least two of the five tracked features changed | 195 | 83.0 |
| Specific anchor was the only one of the five tracked features that changed | 40 | 17.0 |

The five tracked features are specific anchor, primary mechanism, anchor category, primary context, and representational domain. Counts overlap. The multi-feature and specific-anchor-only counts use all five fields. These summaries describe established boundaries; they do not identify causal boundary determinants. The 16 between-segment links are excluded.

Specific-anchor and representational-domain information comes from annotation records. The public input retains a derived anchor-change flag and representational-domain labels, while omitting specific-anchor text and original identifiers.

### D Manual excerpt-traceability audit

| Participant | Total excerpts | Excerpts audited | Observed failures |
|---|---|---|---|
| P01 | 52 | 18 | 0 |
| P02 | 84 | 28 | 0 |
| P03 | 136 | 46 | 0 |
| P04 | 59 | 20 | 0 |
| P05 | 111 | 38 | 0 |
| Total | 442 | 150 | 0 |

Excerpts were randomly sampled without replacement within participants, with proportionally allocated sample sizes. All sampled excerpts were located in their source transcripts and matched the source wording apart from trivial formatting differences. The dataset-wide one-sided 95% upper confidence bound was seven failures among 442 excerpts (1.58%), calculated using the participant-stratified finite-population procedure in Supplement S1.

## Table S2 State-space construction and selected resolutions

### A Parallel-analysis dimensionality retention

| Participant omitted | Events fitted | Retained dimensions | Permutations |
|---|---|---|---|
| None (full sample) | 256 | 13 | 5,000 |
| P01 | 227 | 11 | 5,000 |
| P02 | 200 | 10 | 5,000 |
| P03 | 190 | 11 | 5,000 |
| P04 | 222 | 12 | 5,000 |
| P05 | 185 | 12 | 5,000 |

### B MCA geometries and selected projections

| Space | Role | Selection or retention basis | Raw inertia (%) | Positive corrected inertia (%) | Full-geometry distance r | Selected plane | Mean trustworthiness | Projection distance r |
|---|---|---|---|---|---|---|---|---|
| 4-D | Selected geometry | Separate macro comparison | 21.48 | 72.88 | 0.738 | Dim0 × Dim3 | 0.873 | 0.855 |
| 10-D | Participant-robust | Minimum omission retention | 42.62 | 95.75 | 0.864 | Dim0 × Dim5 | 0.771 | 0.699 |
| 13-D | Full retained | Full-sample parallel analysis | 50.84 | 98.82 | 0.909 | Dim0 × Dim5 | 0.748 | 0.680 |

The 4-D geometry was selected for macro-organization without using transition or successor outcomes as selection criteria. The 10-D participant-robust and 13-D full retained geometries provide higher-dimensional references. All are prefixes of the same MCA fit.

Raw inertia uses total raw inertia; positive corrected inertia uses total positive Benzécri-corrected inertia. Full-geometry r compares distances with the complete 54-D solution. Projection r compares each selected plane with its named geometry. Both are Pearson correlations. Mean trustworthiness compares the projection with that geometry, averaging 5, 10, and 20 neighbors. This differs from the full-geometry fidelity diagnostic in Figure S1B. Complete projection searches and participant-exclusion results are generated by the public reproducibility workflow (Supplement S7).

Table S2 State-space construction and selected resolutions

## C Macro-cluster comparisons

| MCA dimensions | Macro-cluster count | Silhouette | Bootstrap ARI | Minimum cluster n | Minimum participants | Maximum participant share | Mean participant-exclusion ARI | Mean held-out corrected $R^2$ |
|---|---|---|---|---|---|---|---|---|
| 2 | 2 | 0.498 | 0.708 | 76 | 5 | 0.342 | 0.807 | 0.053 |
| 3 | 3 | 0.414 | 0.776 | 45 | 5 | 0.347 | 0.705 | 0.057 |
| 2 | 4 | 0.498 | 0.714 | 1 | 1 | 1.000 | 0.607 | 0.050 |
| 3 | 4 | 0.428 | 0.681 | 1 | 1 | 1.000 | 0.818 | 0.056 |
| 4 | 4 | 0.349 | 0.735 | 17 | 4 | 0.471 | 0.720 | 0.054 |
| 5 | 4 | 0.296 | 0.669 | 17 | 4 | 0.471 | 0.636 | 0.055 |
| 10 | 4 | 0.165 | 0.474 | 17 | 4 | 0.471 | 0.592 | 0.055 |
| 13 | 2 | 0.185 | 0.543 | 84 | 5 | 0.345 | 0.908 | 0.049 |
| 13 | 4 | 0.122 | 0.470 | 30 | 5 | 0.343 | 0.503 | 0.051 |

Each row reports a k-means solution with the specified MCA dimensionality and macro-cluster count. ARI is the adjusted Rand index. These comparisons retain the selected 4-D four-cluster solution, low-dimensional alternatives, its nearest higher-dimensional four-cluster neighbor, and the 10-D and 13-D references. The full two- through 13-dimensional, two- through six-cluster grid is generated by the public reproducibility workflow (Supplement S7).

Eligibility required at least 5% of events and two participants per cluster; the 2-D and 3-D four-cluster solutions failed both criteria. Among eligible four-cluster solutions, the 4-D four-cluster solution had the greatest silhouette and bootstrap ARI. Its participant-exclusion agreement exceeded that of the 10-D and 13-D four-cluster solutions, with similar held-out corrected $R^2$. Higher-dimensional geometries retained more overall distance information (Table S2B). These comparisons support the intended four-cluster macro scale, without establishing universal superiority. Selection excluded transition and successor outcomes.

## D Four-cluster macro descriptions and support

| Cluster | Post hoc description | Events | Participants | Maximum participant share |
|---|---|---|---|---|
| C1 | Social context integration/belonging | 133 | 5 | 0.331 |
| C2 | Goal-directed creative/prospective organization | 59 | 5 | 0.305 |
| C3 | Virtual-digital mediation | 17 | 4 | 0.471 |
| C4 | Physical-spatial place/path organization | 47 | 5 | 0.383 |

Descriptions summarize fitted feature composition and did not define membership. Five of six held-out feature sets aligned with this partition at $q < 0.001$; event-level sentiment did not ($q = 0.114$). These q values adjust six feature sets within the 4-D four-cluster solution; adjustment for the full set of 360 tests across feature sets and clustering solutions gives $q = 0.117$ for event-level sentiment. Tests are described in Supplement S3.

# Table S2 State-space construction and selected resolutions

## E Selected basin solutions and support

| Solution | Fitting space | Event bounds | Actual events per basin | Minimum outgoing | Mean fitting-space silhouette | Lowest basin-mean silhouette | 13-event refit mean ARI |
|---|---|---|---|---|---|---|---|
| 4-D ten-basin | 4-D | 9–47 | 9–47 | 9 | 0.266 | -0.238 | 0.733 |
| 4-D nine-basin | 4-D | 8–49 | 8–49 | 8 | 0.256 | -0.210 | 0.719 |
| 4-D six-basin | 4-D | 8–50 | 17–50 | 16 | 0.230 | 0.146 | 0.662 |
| 2-D nine-basin | 2-D | 8–50 | 12–50 | 12 | 0.381 | 0.190 | 0.575 |
| 2-D eight-basin | 2-D | 8–39 | 15–39 | 15 | 0.332 | 0.207 | 0.521 |

## F Selected 4-D ten-basin composition and transition support

| Basin-state | Modal primary context / primary mechanism | Events | Participants | Maximum participant share | Outgoing transitions | Self-transitions | Self-transition probability |
|---|---|---|---|---|---|---|---|
| B1 | internal simulated / Context integration | 26 | 5 | 0.385 | 26 | 2 | 0.077 |
| B2 | social / Associative learning | 43 | 5 | 0.395 | 42 | 8 | 0.190 |
| B3 | physical / Boundary or landmark encoding; Context integration | 9 | 3 | 0.444 | 9 | 0 | 0.000 |
| B4 | social / Context integration | 47 | 5 | 0.340 | 47 | 15 | 0.319 |
| B5 | social / Context integration | 47 | 5 | 0.362 | 45 | 14 | 0.311 |
| B6 | physical / Goal-directed navigation | 10 | 5 | 0.300 | 10 | 2 | 0.200 |
| B7 | physical / Place coding | 18 | 5 | 0.444 | 18 | 3 | 0.167 |
| B8 | physical / Episodic sequence retrieval | 17 | 5 | 0.647 | 17 | 0 | 0.000 |
| B9 | virtual digital / Context integration | 17 | 4 | 0.471 | 16 | 5 | 0.312 |
| B10 | social / Goal-directed navigation | 22 | 5 | 0.318 | 21 | 3 | 0.143 |

All five solutions partition the same 256 event-states and retain 251 original adjacencies. The 4-D fits use Dim0–Dim3; the 2-D fits use Dim0 × Dim3. Exact source identifiers, in Table S2E order, are D4_K10_L9_U47, D4_K9_L8_U49, D4_K6_L8_U50, D2_K9_L8_U50, and D2_K8_L8_U39. Count bounds are analyst-specified heuristics. Routing and successor outcomes were excluded from selection.

Modal context and mechanism are descriptive; ties are retained. Basin numbers are local to this solution. B3 contains nine events from three participants and has mean fitting-space silhouette −0.238. The 52 self-transitions comprise 20.7% of all 251 adjacencies. Detailed construction settings are in the repository and Supplement S4.

## Table S2 State-space construction and selected resolutions

### G Reserved assignment perturbation diagnostics

| Solution | 13-event refit mean ARI | 13-event refit 10th percentile ARI | Refit support retention | Fixed-coordinate mean ARI | Half/full-start mean ARI |
|---|---|---|---|---|---|
| 4-D ten-basin | 0.733 | 0.575 | 87/90 | 0.874 | 0.925 |
| 4-D nine-basin | 0.719 | 0.598 | 73/90 | 0.852 | 0.959 |
| 4-D six-basin | 0.662 | 0.463 | 88/90 | 0.754 | 0.863 |
| 2-D nine-basin | 0.575 | 0.341 | 89/90 | 0.882 | 0.915 |
| 2-D eight-basin | 0.521 | 0.350 | 90/90 | 0.678 | 0.936 |

Results use 90 reserved omissions of 13 events. Fixed-coordinate and refitted-coordinate trials used 30 starts per fit; the final column compares 15 and 30 starts. All 90 refitted trials completed for every solution. Support retention counts refits with at least eight outgoing observations in every basin-state. These assignment-stability checks differ from the held-out transition-prediction comparison in Table S3F.

## Table S3 Narrative transitions and feature-conditioned routing

### A Macro-cluster transition probability and centroid distance

| Centroid-distance geometry | Role | Observed directed non-self pairs | Pearson r | Exact two-sided p | Spearman ρ |
|---|---|---|---|---|---|
| 4-D | Selected geometry | 12 | -0.465 | 0.208 | -0.573 |
| 10-D | Participant-robust | 12 | -0.451 | 0.167 | -0.573 |
| 13-D | Full retained | 12 | -0.449 | 0.167 | -0.573 |

All 24 centroid-label permutations are enumerated. The 12 observed directed non-self-cluster pairs use fixed four-cluster memberships, with centroid distances calculated in each named geometry.

### B Feature–routing correspondence

| Solution / profile | Feature levels | Basin pairs | Pearson r | Nominal p | Spearman ρ | Nominal p |
|---|---|---|---|---|---|---|
| 4-D ten-basin / primary three families | 22 | 45 | 0.334 | 0.038 | 0.314 | 0.045 |
| 4-D ten-basin / all 11 variables | 65 | 45 | 0.290 | 0.053 | 0.277 | 0.079 |
| 4-D nine-basin / primary three families | 22 | 36 | 0.400 | 0.016 | 0.397 | 0.018 |
| 4-D nine-basin / all 11 variables | 65 | 36 | 0.198 | 0.245 | 0.260 | 0.142 |
| 4-D six-basin / primary three families | 22 | 15 | 0.798 | 0.016 | 0.800 | 0.014 |
| 4-D six-basin / all 11 variables | 65 | 15 | 0.724 | 0.008 | 0.764 | 0.012 |
| 2-D nine-basin / primary three families | 22 | 36 | 0.263 | 0.204 | 0.366 | 0.064 |
| 2-D nine-basin / all 11 variables | 65 | 36 | 0.234 | 0.224 | 0.299 | 0.112 |
| 2-D eight-basin / primary three families | 22 | 28 | 0.596 | 0.028 | 0.680 | 0.008 |
| 2-D eight-basin / all 11 variables | 65 | 28 | 0.591 | 0.013 | 0.649 | 0.006 |

The primary basin-state feature profiles combine category counts for mechanism presence, primary context, and event-level meta-theme. Basin-state feature-profile similarities and routing-profile similarities are calculated using cosine similarity. Mechanism presence combines primary and secondary mechanisms, counting each distinct mechanism once per event. Two-sided tests use 10,000 common basin-label permutations with seed 42 and finite-sample correction. Values are nominal, without correction across solutions or profile definitions; solutions reuse the same events and participants.

For the main three-family comparison, participant omissions yielded r = 0.124–0.504, with two of five nominally significant. The compact main-text summary is Table 3; all-11-variable comparisons are included here.

# Table S3 Narrative transitions and feature-conditioned routing

## C Feature-conditioned routing

| Solution | Feature family | Full-system cosine | Null mean | BH q | Supported-source cosine | Null mean | BH q |
|---|---|---|---|---|---|---|---|
| 4-D ten-basin | Primary context | 0.127 | 0.292 | < 0.001 | 0.511 | 0.501 | 0.547 |
| 4-D ten-basin | Mechanism presence | 0.295 | 0.413 | < 0.001 | 0.537 | 0.596 | 0.403 |
| 4-D ten-basin | Event-level meta-theme | 0.153 | 0.175 | 0.031 | 0.433 | 0.440 | 0.547 |
| 4-D nine-basin | Primary context | 0.175 | 0.346 | < 0.001 | 0.672 | 0.549 | 0.956 |
| 4-D nine-basin | Mechanism presence | 0.392 | 0.460 | 0.008 | 0.581 | 0.635 | 0.452 |
| 4-D nine-basin | Event-level meta-theme | 0.181 | 0.216 | 0.008 | 0.533 | 0.475 | 0.956 |
| 4-D six-basin | Primary context | 0.223 | 0.531 | < 0.001 | 0.754 | 0.641 | 0.973 |
| 4-D six-basin | Mechanism presence | 0.463 | 0.622 | < 0.001 | 0.734 | 0.715 | 0.973 |
| 4-D six-basin | Event-level meta-theme | 0.284 | 0.371 | < 0.001 | 0.594 | 0.532 | 0.973 |
| 2-D nine-basin | Primary context | 0.182 | 0.323 | < 0.001 | 0.461 | 0.528 | 0.269 |
| 2-D nine-basin | Mechanism presence | 0.320 | 0.440 | < 0.001 | 0.466 | 0.618 | 0.003 |
| 2-D nine-basin | Event-level meta-theme | 0.188 | 0.201 | 0.195 | 0.476 | 0.460 | 0.663 |
| 2-D eight-basin | Primary context | 0.250 | 0.375 | < 0.001 | 0.579 | 0.500 | 0.969 |
| 2-D eight-basin | Mechanism presence | 0.341 | 0.488 | < 0.001 | 0.616 | 0.599 | 0.969 |
| 2-D eight-basin | Event-level meta-theme | 0.203 | 0.225 | 0.052 | 0.460 | 0.397 | 0.969 |

Conditions required at least 10 source transitions. Supported-source comparisons required at least three outgoing transitions from the same source basin-state under each compared condition. The retained condition counts were four for context, six for mechanism, and eleven for meta-theme. Lower observed cosine similarity indicates greater differentiation between conditions. BH q values adjust three feature families separately within each comparison and solution; they do not control a study-wide family.

Context and meta-theme partition source transitions; mechanism-presence groups overlap. Support masks were recomputed within each condition-label shuffle. In the main solution, only 3/6 context pairs, 15/15 mechanism pairs, and 35/55 meta-theme pairs share a supported source. The 2-D nine-basin supported-source mechanism result becomes nonsignificant after omitting P01 (q = 0.076) or P05 (q = 0.157).

Raw p values, condition counts, routing matrices, and support breakdowns are generated by the public reproducibility workflow; consolidated results are included in documentation/EXTENDED_RESULTS.md (Supplement S7). Global label shuffles do not preserve participant-specific condition frequencies. Omission analyses reapply eligibility, so retained conditions can change.

## Table S3 Narrative transitions and feature-conditioned routing

### D Consecutive event-state distances in the 4-D geometry

| Participant | Event-state adjacencies | Observed mean distance | Expected mean under shuffled order | Reduction (%) |
|---|---|---|---|---|
| P01 | 28 | 0.994 | 1.181 | 15.8 |
| P02 | 55 | 1.252 | 1.328 | 5.7 |
| P03 | 65 | 0.982 | 1.247 | 21.3 |
| P04 | 33 | 1.269 | 1.374 | 7.7 |
| P05 | 70 | 1.046 | 1.089 | 4.0 |
| Pooled | 251 | 1.098 | 1.230 | 10.7 |

The pooled mean weights each adjacency equally. The null shuffles interior event order within participants with first and last event-states fixed. The pooled two-sided test measures absolute departure from the exact null-expected mean. Its p value of $1 \times 10^{-6}$ is the plus-one resolution limit, with zero exceedances in 1,000,000 shuffles. Participant rows are descriptive. This test is separate from the six-test adjustment in Table S3E. Figure 2B displays these distances.

### E Displacement field and macro-to-next-basin dependence

| Analysis or basin solution | Statistic | Observed | Null mean | Observed minus null mean | Upper-tail p | Holm p across six tests |
|---|---|---|---|---|---|---|
| Dim0 × Dim3 displacement field | Weighted squared field departure | 0.117 | 0.081 | 0.037 | 0.0014 | 0.0070 |
| 4-D ten-basin | Conditional mutual information | 0.298 | 0.241 | 0.057 | 0.0139 | 0.0417 |
| 4-D nine-basin | Conditional mutual information | 0.278 | 0.214 | 0.064 | 0.0063 | 0.0252 |
| 4-D six-basin | Conditional mutual information | 0.230 | 0.150 | 0.081 | 0.0004 | 0.0024 |
| 2-D nine-basin | Conditional mutual information | 0.256 | 0.220 | 0.036 | 0.0710 | 0.1419 |
| 2-D eight-basin | Conditional mutual information | 0.237 | 0.209 | 0.029 | 0.1111 | 0.1419 |

All six statistics use the same 1,000,000 endpoint-fixed within-participant interior-order shuffles. The field statistic measures source-count-weighted squared departure from the exact expected field in squared raw MCA-coordinate units. It tests the pooled Dim0 × Dim3 field globally, not individual arrows or conditioned fields (Figure 2C–D).

The other rows measure mutual information between current macro-cluster and next basin-state conditional on participant, in nats. The macro partition is fixed. Raw conditional mutual information has positive finite-sample bias; null means and excesses are reported. Holm correction covers these six tests.

### F Held-out prediction gain from current basin-state identity

| Basin solution | Mean log-score gain (nats per adjacency) | Participants with positive gain | Adjacencies meeting support heuristic | Zero-training-support adjacencies |
|---|---|---|---|---|
| 4-D ten-basin | -0.097 | 0/5 | 198/251 (78.9%) | 4 |
| 4-D nine-basin | -0.115 | 0/5 | 225/251 (89.6%) | 4 |
| 4-D six-basin | -0.031 | 0/5 | 246/251 (98.0%) | 3 |
| 2-D nine-basin | -0.117 | 0/5 | 203/251 (80.9%) | 2 |
| 2-D eight-basin | -0.108 | 0/5 | 207/251 (82.5%) | 3 |

Each participant's next basin-state was predicted using transition probabilities estimated from the other four participants. Gain compares current macro-cluster plus basin identity with current macro-cluster alone; positive values favor adding basin identity. Coordinates and memberships remain fixed. The support heuristic requires at least eight training adjacencies from three training participants for the source macro-cluster–basin combination. Zero support falls back to the macro-only predictor. All five mean gains were negative; no significance test was applied.

# Table S4 Successor-profile correspondence and controls

## A Representational–successor correspondence at γ = 0.90

| Basin solution | Basin pairs | 4-D r / nominal p | 10-D r / nominal p | 13-D r / nominal p |
|---|---|---|---|---|
| 4-D ten-basin | 45 | 0.427 / 0.080 | 0.432 / 0.100 | 0.431 / 0.101 |
| 4-D nine-basin | 36 | 0.415 / 0.110 | 0.401 / 0.143 | 0.400 / 0.146 |
| 4-D six-basin | 15 | 0.832 / 0.005 | 0.803 / 0.011 | 0.802 / 0.011 |
| 2-D nine-basin | 36 | 0.455 / 0.035 | 0.453 / 0.042 | 0.425 / 0.066 |
| 2-D eight-basin | 28 | 0.684 / 0.001 | 0.662 / 0.002 | 0.655 / 0.003 |

Complete successor profiles are rows of $M = (I - \gamma P)^{-1}$, including current-state identity and expected discounted subsequent occupancy. Correlations compare Euclidean centroid distances in each named geometry with distances between complete successor profiles. All solutions use 251 original event adjacencies; their prefixes indicate fitting space. Tests use two-sided common basin-label permutations. These dependent comparisons are not independent replications.

The main discount is $\gamma = 0.90$. Results for the full $\gamma = 0.70$, 0.80, 0.90, and 0.95 sweep are provided in the "Successor discount geometry sweep" table in documentation/EXTENDED_RESULTS.md within the public reproducibility package (Supplement S7). Nominal p values are retained here; cross-solution inference is reported in Tables S4G–H.

## B Two-dimensional alignment for the displayed solutions

| Solution | Successor PC1 fraction | Successor PC2 fraction | Cumulative fraction | Procrustes disparity | Lower-tail p | Permutations |
|---|---|---|---|---|---|---|
| 4-D ten-basin | 0.200 | 0.161 | 0.361 | 0.992 | 0.193 | 10,000 |
| 2-D nine-basin | 0.206 | 0.185 | 0.391 | 0.943 | 0.209 | 10,000 |

Variance entries are fractions. Procrustes compares Dim0 × Dim3 with the first two successor principal components using a lower-tail test (Figures 5A and S5A–C). It is distinct from complete-profile correspondence. Results for all five solutions are provided in the "Procrustes all solutions" table in documentation/EXTENDED_RESULTS.md within the public reproducibility package (Supplement S7). The main projection-restricted distance comparison gave $r = 0.512$, nominal $p = 0.027$, and is secondary to the complete-profile result.

## C Participant-exclusion summary

| Solution / unit | Omissions | Fixed-centroid r range | Fixed-centroid p < 0.05 | Recomputed-centroid r range | Recomputed-centroid p < 0.05 |
|---|---|---|---|---|---|
| 4-D ten-basin / event-state adjacency | 5 | 0.246–0.492 | 1/5 | 0.277–0.583 | 1/5 |
| 4-D ten-basin / basin-visit adjacency | 5 | 0.430–0.615 | 5/5 | 0.428–0.622 | 5/5 |
| 2-D nine-basin / event-state adjacency | 5 | 0.296–0.530 | 3/5 | 0.294–0.544 | 3/5 |
| 2-D nine-basin / basin-visit adjacency | 5 | 0.410–0.551 | 5/5 | 0.393–0.542 | 5/5 |

Each omission rebuilds P and M while retaining original MCA coordinates and basin memberships. Centroids are either fixed or recomputed from retained events. All correlations remain positive. Ranges and support counts summarize the five omissions at $\gamma = 0.90$ using 4-D distances; full-sample rows are excluded. These are participant-influence checks, not participant-specific systems or full-pipeline refits. Omission-specific results and nominal p values are provided in the "Participant exclusion complete" table in documentation/EXTENDED_RESULTS.md within the public reproducibility package; full-precision outputs are generated by the workflow (Supplement S7).

## Table S4 Successor-profile correspondence and controls

### D Event-adjacency and basin-visit comparison

| Solution | Event self-transitions | Visit adjacencies | Event r / p | Visit r / p | Δr |
|---|---|---|---|---|---|
| 4-D ten-basin | 52 | 199 | 0.427 / 0.080 | 0.585 / < 0.001 | 0.158 |
| 4-D nine-basin | 52 | 199 | 0.415 / 0.110 | 0.651 / < 0.001 | 0.236 |
| 4-D six-basin | 72 | 179 | 0.832 / 0.005 | 0.803 / 0.002 | -0.028 |
| 2-D nine-basin | 49 | 202 | 0.455 / 0.035 | 0.552 / 0.002 | 0.097 |
| 2-D eight-basin | 48 | 203 | 0.684 / 0.001 | 0.691 / < 0.001 | 0.007 |

Basin visits collapse maximal consecutive runs within one basin-state. All event comparisons use 251 adjacencies; visit adjacencies have no self-transitions. The main solution has 204 visits and 199 visit adjacencies; the 2-D nine-basin solution has 207 and 202. One discounted step becomes one visit-to-visit transition. All comparisons use 4-D centroids and $\gamma = 0.90$. Δr is descriptive and was calculated before rounding; it does not test differences between correlations. Figure S6A–B shows the main and 2-D nine-basin visit comparisons.

### E Narrated and reversed order relative to shuffling

| Solution | Narrated r | Narrated order p | Reversed r | Reversed order p | Shuffle central 95% |
|---|---|---|---|---|---|
| 4-D ten-basin | 0.427 | 0.062 | 0.432 | 0.059 | -0.256 to 0.504 |
| 4-D nine-basin | 0.415 | 0.113 | 0.399 | 0.126 | -0.243 to 0.561 |
| 4-D six-basin | 0.832 | 0.003 | 0.800 | 0.008 | -0.402 to 0.741 |
| 2-D nine-basin | 0.455 | 0.071 | 0.499 | 0.041 | -0.249 to 0.530 |
| 2-D eight-basin | 0.684 | 0.002 | 0.672 | 0.002 | -0.325 to 0.518 |

Upper-tail order tests rebuild P and M after independently shuffling each participant's event order, allowing endpoints to change. This differs from the endpoint-fixed null in Tables S3D–E. Distances use 4-D centroids at $\gamma = 0.90$. The central 95% is the shuffle-null range, not uncertainty around observed r. Narrated and reversed tests can both reject this null; these results do not establish forward-direction specificity or a difference between correlations. Label-permutation and order-shuffle results are summarized in the "Narrated reversed order complete" table in documentation/EXTENDED_RESULTS.md within the public reproducibility package; detailed outputs are generated by the workflow (Supplement S7; Figure S6D).

### F Profile-definition controls

| Solution | P r / p | I + 0.90P r / p | M r / p | M − I r / p |
|---|---|---|---|---|
| 4-D ten-basin | 0.516 / 0.028 | 0.377 / 0.137 | 0.427 / 0.080 | 0.513 / 0.029 |
| 4-D nine-basin | 0.539 / 0.008 | 0.349 / 0.190 | 0.415 / 0.110 | 0.578 / 0.005 |
| 4-D six-basin | 0.834 / 0.008 | 0.798 / 0.016 | 0.832 / 0.005 | 0.871 / 0.006 |
| 2-D nine-basin | 0.408 / 0.068 | 0.453 / 0.039 | 0.455 / 0.035 | 0.436 / 0.061 |
| 2-D eight-basin | 0.749 / 0.002 | 0.690 / < 0.001 | 0.684 / 0.001 | 0.741 / 0.001 |

Values compare 4-D centroid distances with distances between the indicated profiles at $\gamma = 0.90$. Rows of M are complete successor profiles, including current-state identity and expected discounted subsequent occupancy. Rows of M − I are identity-removed successor profiles. The P comparison uses Euclidean distances between routing profiles. Table S3B instead correlates pairwise cosine similarities between basin-state feature profiles with pairwise cosine similarities between routing profiles. Nominal p values use separate tests and do not establish differences between correlations or the necessity of accumulated multistep structure. No adjustment across these profile definitions is implied (Figure S6C).

## Table S4 Successor-profile correspondence and controls

### G Across-solution partial-conjunction tests

| Requirement within each basin solution | Required number of basin solutions | Partial-conjunction p |
|---|---|---|
| 4-D distances | At least two | 0.022 |
| 4-D distances | At least three | 0.105 |
| Every distance space (4-D, 10-D, 13-D) | At least two | 0.044 |
| Every distance space (4-D, 10-D, 13-D) | At least three | 0.198 |

Bonferroni partial conjunction permits dependent component tests. Values use exact label enumeration for the 4-D six-basin and 2-D eight-basin solutions, and Monte Carlo tests for larger solutions. Fixed-sequence tests across counts support a lower bound of two solutions. The original Monte Carlo versions round to the same values. These claims concern the existing solutions and basin-label nulls.

### H Solution identification with Holm correction

| Basin solution | 4-D distances, adjusted p | Every distance space, adjusted p |
|---|---|---|
| 4-D ten-basin | 0.160 | 0.203 |
| 4-D nine-basin | 0.160 | 0.203 |
| 4-D six-basin | 0.022 | 0.044 |
| 2-D nine-basin | 0.105 | 0.198 |
| 2-D eight-basin | 0.007 | 0.013 |

Holm correction covers five solutions within each column. The all-geometry input is each solution's largest component p across 4-D, 10-D, and 13-D distances. The 4-D six-basin and 2-D eight-basin solutions retain support. These nested assessments reuse the same events. Table 3 and Figure 5B report the all-geometry result.

### I Joint narrated-order test across five solutions

| Centroid distance space | Observed median r | Shuffle-null median | Joint upper-tail p |
|---|---|---|---|
| 4-D | 0.455 | 0.161 | 0.037 |
| 10-D | 0.453 | 0.157 | 0.041 |
| 13-D | 0.431 | 0.155 | 0.054 |

The statistic is the median correlation across five solutions, recalculated after each of 10,000 shared within-participant event-order shuffles. Endpoints may change; all P and M matrices are rebuilt. Complete profiles use $\gamma = 0.90$. The primary distance geometry is 4-D.